\documentclass[conference]{IEEEtran}
\IEEEoverridecommandlockouts
\usepackage{cite}
\usepackage{amsmath,amssymb,amsfonts}
\usepackage{graphicx}
\usepackage{textcomp}
\usepackage{xcolor}

\usepackage[utf8]{inputenc} 
\usepackage[T1]{fontenc}    
\usepackage{hyperref}       
\usepackage{url}            
\usepackage{booktabs}       
\usepackage{amsfonts}       
\usepackage{nicefrac}       
\usepackage{microtype}      
\usepackage{xcolor}         
\usepackage{amsmath}  

\usepackage{algorithm}
\usepackage{algpseudocode}

\usepackage{graphicx}
\usepackage{wrapfig}

\usepackage{caption}
\usepackage{subcaption}

\def\BibTeX{{\rm B\kern-.05em{\sc i\kern-.025em b}\kern-.08em
    T\kern-.1667em\lower.7ex\hbox{E}\kern-.125emX}}
\begin{document}



\title{Using Cooperative Game Theory to Prune Neural Networks}

%



\author{

\IEEEauthorblockN{Mauricio Diaz-Ortiz Jr}
\IEEEauthorblockA{\textit{Dodners Institute and Radboud University}
}

\IEEEauthorblockN{ Benjamin Kempinski }
\IEEEauthorblockA{\textit{Dodners Institute and Radboud University}
}

\IEEEauthorblockN{Daphne Cornelisse}
\IEEEauthorblockA{\textit{New York University}
}

\IEEEauthorblockN{Yoram Bachrach}
\IEEEauthorblockA{\textit{Google DeepMind}
}

\IEEEauthorblockN{Tal Kachman}
\IEEEauthorblockA{\textit{Dodners Institute and Radboud University}
}
}

\maketitle

\begin{abstract}
    We show how solution concepts from cooperative game theory can be used to tackle the problem of pruning neural networks.
    The ever-growing size of deep neural networks (DNNs) increases their performance, but also their computational requirements. We introduce a method called Game Theory Assisted Pruning (GTAP), which reduces the neural network's size while preserving its predictive accuracy. GTAP is based on eliminating neurons in the network based on an estimation of their joint impact on the prediction quality through game theoretic solutions. Specifically, we use a power index akin to the Shapley value or Banzhaf index, tailored using a procedure similar to Dropout (commonly used to tackle overfitting problems in machine learning). 
    Empirical evaluation of both feedforward networks and convolutional neural networks shows that this method outperforms existing approaches in the achieved tradeoff between the number of parameters and model accuracy. 
\end{abstract}

\maketitle

\section{Introduction}
\begin{figure}[h!t]
    \centering
    \begin{subfigure}[b]{0.43\textwidth}
         \centering
         \includegraphics[width=1\textwidth]{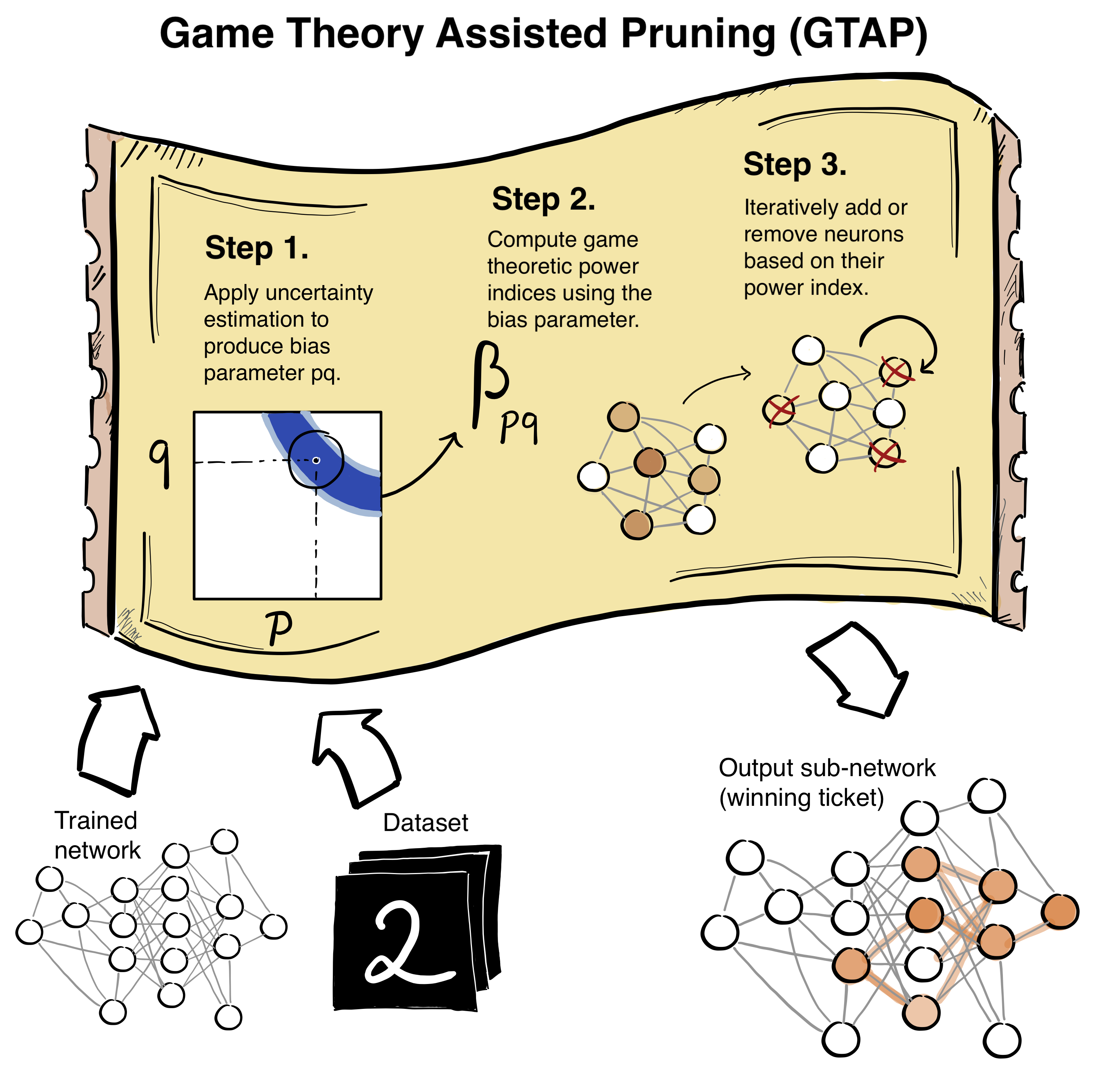}
         \caption{}\label{fig:overall_approach}
     \end{subfigure}
     \begin{subfigure}[b]{0.42\textwidth}
         \centering
         \includegraphics[width=1\textwidth]{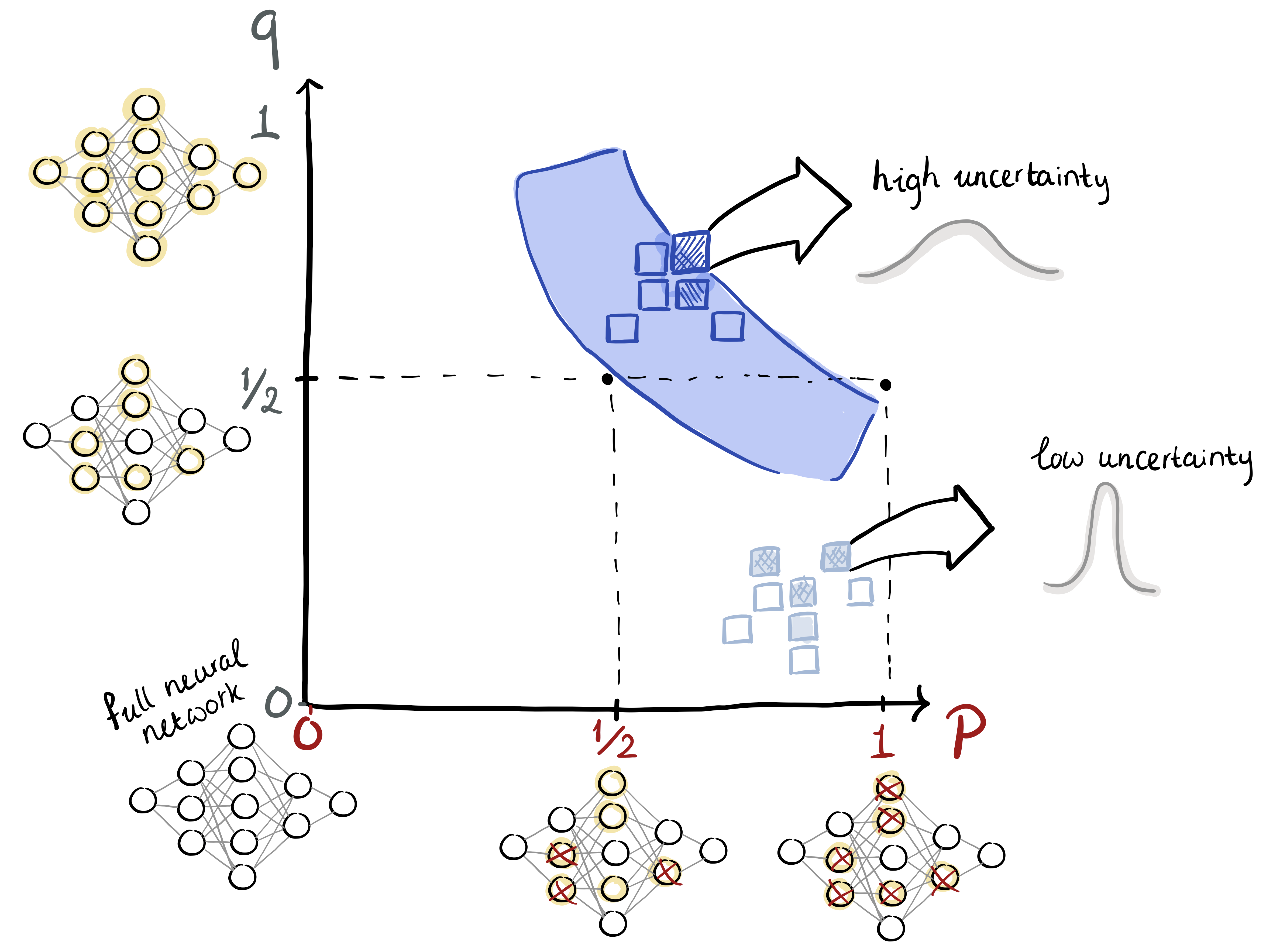}
         \caption{}\label{fig:Uncert_band}
     \end{subfigure}
    \caption{Top: The steps of our proposed procedure GTAP for pruning neural networks. GTAP takes the data and model and calculates uncertainty bands using a Dropout-based process. It then computes game theoretic power indices for neurons, and uses them to decide which neurons to prune. Bottom: sketch of the uncertainty band estimation.}
\end{figure}\label{fig:paper_flow}

    Many of the successes in machine learning in the last two decades are attributed to deep neural networks (DNNs), from natural language processing~\cite{otter2020survey,brown2020language}, through vision~\cite{voulodimos2018deep,dosovitskiy2020image} to fundamental problems in science~\cite{jumper2021highly,reichstein2019deep}. The success of such models depends on the vast number of parameters they contain and the complexity they exhibit, where modern networks can easily reach more than 500 billion parameters~\cite{smith2022using}, requiring a vast amount of compute. As a result of their number of parameters, such large models consume a significant amount of power during inference and prediction and cannot be deployed on resource-constrained edge devices~\cite{al2015efficient,murshed2021machine,metz2021gradients}. 
       
    The neural network's performance is attained through a joint computation using many individual neurons. Cooperative game theory offers tools for analyzing the interactions between different team members and how the overall performance is impacted by the individual members. Can we leverage these tools to reduce the computational requirements of neural networks, without degrading their predictive performance?

    The most direct approach is to reduce the size of the network, in terms of either the number of neurons or the number of connections between them, referred to as {\em pruning}~\cite{blalock2020state}. Several pruning methodologies have been proposed, including stochastic regularization~\cite{lecun1990optimal, Hassibi1993}, co-adaptation~\cite{han2015learning} and activation-based metrics~\cite{Hu2016, Zhao2022}. In some cases, there may even exist sub-networks whose prediction accuracy {\em exceeds} that of the original full network, sometimes referred to as ``winning lottery tickets''~\cite{frankle2019lottery}. However, providing methods for finding such winning tickets is a significant algorithmic challenge ~\cite{Chen2020, Malach2020}.

    Rather than enumerating the underlying space of sub-networks or relying on simple heuristics, we propose seeking high performing sub-networks (winning tickets) by viewing the neurons of a network as agents playing a cooperative game, where the neurons work together to maximize network performance. This enables us to use game theoretic tools that measure each participant's contribution to the collective goal~\cite{Shapley1953, BanzhafIII1964, branzei2008models, chalkiadakis2011computational,cornelisse2022neural}, with the added benefit of interpretability. Further, these cooperative game principles are model agnostic and thus can be reformulated in other learning problems.


\textbf{Our Contribution}:
We propose Game Theory Assisted Pruning (GTAP), a method for pruning neural networks based on cooperative game theory. We define a cooperative game, which views the individual neurons of the trained network as agents working in teams, aiming to  produce a highly accurate predictive model. 

Under this view, the value of every subset (coalition) of neurons is the quality of the prediction using a network that uses solely these neurons (with all other activations masked out). We then estimate the relative impact of each neuron on the performance of the entire network using solution concepts from cooperative game theory. Given the relative impact estimate of each neuron we construct the sub-network by retaining only the high impact neurons, or adding neurons gradually in decreasing order of impact. 

To determine the relative impact of neurons, we use power indices~\cite{Chalkiadakis2011} --- existing game theoretic solutions designed to estimate the impact that individual members of a team have on the overall team performance, such as the Shapley value~\cite{Shapley1953} and the Banzhaf index~\cite{BanzhafIII1964}. We also propose a parameterized version of the Banzhaf index, called $\beta_d$, where $d$ is a parameter reflecting the predicted proportion of neurons required to get a good prediction. To select the parameter $d$, we apply an {\em uncertainty estimation process}, akin to the Dropout procedure~\cite{srivastava2014dropout} commonly used to reduce model over-fitting in machine learning. Our uncertainty estimation procedure considers randomly eliminating neurons in the trained model, and attempts to characterize the network size where we transition from being relatively certain about making a good prediction to being uncertain about our ability to have a high performing model. 

The entire GTAP method is illustrated in Figure~\ref{fig:paper_flow} (with the uncertainty estimation process shown on the bottom). As the figure illustrates, we use the uncertainty estimates regarding the model's quality to select an appropriate parameter $d$. Given this parameter, we select the appropriate solution from cooperative game theory, and use the power index $\beta_d$ to measure the relative impact of individual neurons, which in turn determines which neurons we decide to prune. The remaining neurons comprise the ``winning-ticket'' -- a small sub-network, whose performance is close to that of the original model. 

We empirically evaluate our framework by pruning several prominent neural network architectures. For image classification we consider the convolutional neural network LeNet5~\cite{lecun1998lenet} on MNIST~\cite{lecun1998lenet}, and for for natural language processing tasks, we consider a feedforward model on news topic classification~\cite{TwitterFinancialNewsTopic2022} and emotion classification in social media texts~\cite{saravia-etal-2018-carer}. We also consider the issue of scaling up to large neural networks, reporting results for the AlexNet~\cite{krizhevsky2012alexnet} architecture on Tiny ImageNet~\cite{le2015tiny}. We show that our game theory pruning methods can outperform existing pruning baselines.



%
%

 
\section{Preliminaries and Related Work}
\label{l_sect_prelim}

We provide a brief review of several topics relating to the key building blocks for our approach: power indices from cooperative game theory~\cite{chalkiadakis2011computational}, and Dropout~\cite{srivastava2014dropout} from machine learning (and its relation to uncertainty estimation~\cite{gal2016uncertainty}.

\subsection{Cooperative Game Theory} \label{sec:coop_game}
Cooperative game theory studies strategic teamwork between self-interested agents, determining which teams (coalitions) would form, and how to allocate the joint gains from cooperation between the individual agents forming the team. 
A (transferable-utility) coalitional game $G$ is defined by a set $A=\{a_1,a_2,\ldots,a_n\}$ of $n$ agents, and a characteristic function $\nu:2^A\rightarrow\mathbb{R}$ which maps a coalition (subset) of players $C\subseteq N$ to a real value $\nu(C)$ which reflects the team's utility or performance. The {\it marginal contribution} of agent $a_i$ to the coalition $C$ reflects the increase in utility the coalition gains when $a_i$ joins it, defined as $m(a_i,C) = \nu(C\cup\{a_i\})-\nu(C)$. We can define a similar concept for {\it permutations} of agents.  We denote the predecessors of $a_i\in A$ in the the permutation $\pi$ as $b(a_i,\pi)$, i.e. the agents appearing before $i$ in the permutation $\pi$. We denote the set of all permutations over the $n$ agents as $\Pi$. The  marginal contribution of player $a_i$ in the permutation $\pi$ is the increase in utility $a_i$ provides when joining the team of agents appearing before them in the permutation,  $m(a_i,\pi)=\nu(b(a_i,\pi)\cup \{a_i\})-\nu(b(a_i,\pi))$. 

One line of work on cooperative game theory seeks to determine the relative impact of agents working in teams, or the fair share of utility they are entitled to. Several such {\it power indices} have been proposed, such as the Shapley value~\cite{Shapley1953} and the Banzhaf index~\cite{BanzhafIII1964}, reflecting different fairness axioms~\cite{dubey1975uniqueness, dubey1979mathematical,Chalkiadakis2011}. 

There is large body of earlier work dealing with applications of power indices~\cite{winter2002shapley,aziz2013shapley,tarashev2016risk,aziz2016study,algaba2019handbook,yazdanpanah2019strategic,bachrach2020negotiating,levinger2020computing,skibski2020signed,yan2021if}. There are also many such applications in machine learning~\cite{rozemberczki2022shapley}, ranging from determining the relative impact of features or data on a model's predictions~\cite{datta2016algorithmic,lundberg2017unified,ghorbani2019data}, through vocabulary selection~\cite{patel2021game}, to federated learning~\cite{nagalapatti2021game} and allocating rewards in reinforcement learning~\cite{banarse2018body,han2022stable}.

The {\bf Shapley value}~\cite{Shapley1953} of agent $i$ is defined as their marginal contribution, averaged across all permutations:

\begin{equation} \label{eq:shapley}
	\phi(i) = \frac{1}{N!}\sum_{\pi\in\prod} \nu (b(a_i,\pi))\cup\{a_i\})-\nu (b(a_i,\pi)) 
\end{equation}

Similarly, agent $i$'s {\bf Banzhaf index}~\cite{BanzhafIII1964}, denoted $\beta(\nu)=(\beta_1,..,\beta_N)$, is their marginal contribution averaged over all possible coalitions that do not contain that agent:
\begin{equation} \label{eq:Banzhaf_def}
    \beta(i)  = \frac{1}{2^{n-1}}\sum _{C\subset A|a_i\notin C} \nu (C\cup\{a_i\})-\nu(C)) 
\end{equation}
Consider an agent $a_i$ and random process for generating a coalition $C$ where we make a Bernoulli trial for each agent $a_j \neq a_i$, where with probability $\frac{1}{2}$ they are included in $C$ and with probability $\frac{1}{2}$ they are excluded from it. Denote this random coalition construction process as $C \sim R_{\frac{1}{2}}$. Consider the above definition for the Banzhaf index $\beta(i)$. As each sub-coalition $C \subset A \setminus \{ a_i \}$ occurs exactly once in the sum for $\beta(i)$, we can rewrite this sum as an expectation: $\beta(i) = E_{C \sim R_{\frac{1}{2}}} [\nu (C\cup\{a_i\})-\nu(C))]$.  Hence, the Banzhaf index of agent $i$ can be viewed as the expected increase in utility that is achieved  under uncertainty about the participation of other agents (i.e. a fair coin is flipped to determine whether each of the other agents participates or not). \footnote{Similarly, the Shapley value reflects the expected utility increase that agent $i$ achieves under uncertainty about the ordering in which agents enter.}

We define a slightly different process $R_d$ for building the coalition $C$, where each agent $a_j \neq a_i$ is included with probability $d$ and excluded with probability $1-d$, obtaining a parametrized index: 
$$\beta_d(i) = E_{C \sim R_d} [\nu (C\cup \{ a_i \})-\nu(C))]$$
We refer to the parametrized version $\beta_d$ as the {\it biased power index} due to the biased sampling. This biased power index plays a key role in how we decide on neurons to prune.

\subsection{Uncertainty Quantification via Dropout}
Dropout~\cite{srivastava2014dropout} is a method proposed for avoiding overfitting~\cite{ying2019overview} when training neural network based machine learning models. In this process, neurons are randomly masked out (``dropped-out'') during training, so as to encourage producing a model that distributes knowledge across many neurons. 

In contrast to Bayesian and probabilistic methods, neural models have no direct measure of uncertainty~\cite{blundell2015weight}.  Recent work has shown that stochastic regularization such as Dropout can act as approximate Bayesian inference to estimate neural model uncertainty~\cite{kendall2017uncertainties,gal2015bayesian}, via a procedure called Monte Carlo Dropout\cite{gal2016uncertainty}. 
In a vanilla Monte Carlo Dropout trial~\cite{JMLR:v15:srivastava14a,gal2016uncertainty} one randomly eliminates each neuron with probability $p$. Each such trial yields a different subset of retained neurons, resulting in a different model output. The variance in the model outputs across many trials can serve as an uncertainty estimate. We explore how quantifying the uncertainty using such stochastic methods can offer insight into the likely size of a ``winning ticket''.

\section{Methods}
\label{sect:methods}

We propose a pruning approach for taking a trained neural network model, and selecting a ``wining ticket'' --- a sub-network of neurons which retains a high predictive performance despite having a low size. Finding such a winning ticket can be computationally prohibitive in large models~\cite{Malach2020}, and in the worst case entails an exponential search problem over the possible sub-network space~\cite{tan2019efficientnet}. 

Our approach for neural network pruning consists of two steps. First, we identify a bias parameter $d$, reflecting a good estimate regarding the size of the sub-network (estimation of the size of the winning ticket). To select the parameter $d$, we use a method which examines the neural network to be pruned and constructs ``uncertainty bands'' that reflect the entropy of the model output under random processes for dropping out neurons. The uncertainty band based process for selecting the parameter $d$ is described in Section~\ref{sect:methods_uncertainty_bands}. Secondly, we define a cooperative game where we view the neurons as agents working in teams to improve the model's accuracy. We then use power indices from cooperative game theory to analyze the neurons in the network and determine their relative impact on the quality of the model, and construct a sub-network from the neurons with the highest estimated impact. We consider both standard power indices such as the Shapley value or Banzhaf index, as well as index $\beta_d$, with the parameter $d$ selected in the first step. This step is described in Section~\ref{sect:methods_GTAP}. The overall process is shown in Figure~\ref{fig:overall_approach}. 


\subsection{Uncertainty Bands} 
\label{sect:methods_uncertainty_bands}

Our goal in the uncertainty band process described here is to select a parameter $d$ which reflects an estimate of the size of the ``winning ticket''. Our pruning method uses this parameter to choose an appropriate power index $\beta_d$ (parametrized by $d$) which is then applied to select the neurons to retain when pruning the network. 

In the full neural network of a trained model, there is significant redundancy of computation, so even when eliminating many random neurons we can be fairly certain of getting the correct output. In contrast, consider a {\em minimal} sub-network where each neuron is critical to getting an accurate prediction on many inputs.     
\footnote{For example, we desire correct predictions on many instances or structures at the minimal possible compute cost~\cite{justus2018predicting,hodl2023explainability,yuksel2023selformer}, or alternatively for deep reinforcement learning settings, we desire correct decisions in many possible states~\cite{bakker2019rlboa,lorraine2021lyapunov,ma2020feudal,lorraine2021using,bhalla2019training,hogewind2022safe,yang2020efficient,kramar2022negotiation,xu2022performance} while minimizing the neural network size.
}
For this sub-network, when we eliminate neurons we cannot be certain of the output of the model. Hence the uncertainty regarding the model output is indicative of the winning ticket size. 

Consider a trained network comprised of $n$ neurons, and a fixed sub-network $S$ of size $m \cdot n$, where $0 \leq m \leq 1$ is the proportion of neurons in the sub-network (so the fraction of removed neurons is $q=1-m$). Suppose we take the sub-network $S$ and apply a dropout procedure which removes each of its neurons with probability $p$ (i.e. independently for each neuron in $S$ we flip a biased coin to decide whether to keep it or not). Suppose we then take the remaining sub-network, diluted by these two steps (the first eliminating a fraction $q$ of the neuron, and the second eliminating in expectation a fraction $p$ of the surviving neurons), and apply it to a random dataset instance. How uncertain are we of the model output? To estimate this uncertainty, we can apply this two-phased process many times, and examine the variance in the resulting outputs. 

As discussed earlier, a high uncertainty under parameters $p,q$ of the process indicates a good candidate for proportion of eliminated neurons in the ``winning ticket''). \footnote{The parameters $p$ and $q$ both relate to a dilution in the neurons of the network but play a different role. The random selection of a sub-network of size $mn=(1-q)n$ neurons yields a sub-network where a fraction of {\em exactly} $q$ of the neurons are eliminated. The $p$ dropout phase takes the $(1-q)n$ surviving neurons, and eliminates a fraction $p$ of them {\em in expectation} --- each neuron is independently eliminated with probability $p$, resulting in a Gaussian distribution for the size of the retained network.}

Based on this intuition, we define a Dropout-based random algorithm which takes in two parameters $p,q$ and produces an uncertainty estimate for these parameters. The procedure is called MCUE$(p,q)$, for Monte-Carlo Uncertainty Estimation. MCUE performs simulations for eliminating neurons in a neural network of $n$ neurons, based on the parameters $p,q$. 
Given a model $\mathcal{N}$ with $n$ neurons, we denote by $S_i \sim f_{1-q} (\mathcal{N})$ the process of sampling a sub-network (subset) $S_i \subseteq \mathcal{N}$ of neurons of size $(1-q) \cdot n$ uniformly at random from the set of all such sub-networks. Further, given a dataset $\mathcal{D}$ we denote by $d \sim U(\mathcal{D})$ sampling an instance $d \in \mathcal{D}$ uniformly at random from all the instances in $\mathcal{D}$. 

MCUE samples sub-networks $S_1, S_2, \ldots, S_k$ where each $S_i \sim f_{1-q}(\mathcal{N})$.
For each sub-network $S_i$ it applies a random Dropout process, where each neuron in the sub-network $S_i$ is retained with probability $1-p$ and eliminated with probability $p$, yielding a sub-network $S'_i$ (we denote the remaining sub-network after removing the neurons in the set $R$ as $\mathcal{N} \setminus R$). Finally, it applies $S'_i$ on a random instance $d_i \sim U(\mathcal{D})$ to obtain the sub-network model prediction $S'_i(d_i)$. The sequence $ t = (t_1, \ldots, t_k) =  (S'_1(d_1), S'_2(d_2), \ldots, S'_k(d_k) )$ is a random sample of model predictions under a two-phased elimination process, with parameters $p,q$. As a proxy for the uncertainty of the resulting model prediction, we use an estimator of the sample variance $Var(t)$. This procedure is given in Algorithm \ref{alg:MCUE}.

\begin{algorithm}
\begin{algorithmic}[1]
\caption{Monte Carlo Uncertainty Estimation (MCUE)}
\label{alg:MCUE}
\Procedure{MCUE}{$p, q, \mathcal{N}, \mathcal{D}$}
    \For{$i=1$ to $k$}
        \State $S_i \sim f_{1-q}(\mathcal{N})$ \Comment{Random sub-network of size $(1-q) \cdot n$}
        \State $R \gets \emptyset$ \Comment{Neurons to drop out}
        \For{$j=1$ to $n$}
            \State $r \gets \mathcal{U}(0, 1)$
            \If{$r < p$} \Comment{Remove neuron with probability $p$}
                \State $R \gets R \cup \{ j \}$
            \EndIf
        \EndFor
        \State $S'_i \gets S_i \setminus R$
        \State $d_i \sim U(\mathcal{D})$
        \State $t_i \gets S'_i(d_i)$
    \EndFor
    \State \Return $Var(t_1, \ldots, t_k)$
\EndProcedure
\end{algorithmic}
\end{algorithm}

MCUE(p,q) is akin to Monte-Carlo Dropout~\cite{gal2015bayesian,gal2016uncertainty}).
To select the parameter $d$, we examine the configuration $(p^*,q^*)$ of maximal uncertainty 
along the diagonal where $p = q$, i.e. $t^* = \arg \max_{p} \text{MCUE}(p,p)$. 
As the bias parameter $d$ for $\beta_d$, we use the value of $d=1-t^*$. Viewing the ratio of $p$ and $q$ as a bias-variance trade-off, the distribution on the diagonal is a plausible indicator for the optimal network size. \footnote{We provide some empirical support for this hypothesis in Appendix C.}





\subsection{Game Theoretic Pruning}
\label{sect:methods_GTAP}

We define a characteristic function where the set of agents are neurons in the full model to be pruned, and the value $\nu(C)$ of a ``team'' $C$ of neurons is the model's performance when retaining only these neurons. 
We consider classification tasks, so $\nu(C)$ denotes the model's test set accuracy using only the neurons in $C$. 

\paragraph{Estimating Power Indices: } Despite their theoretical appeal for measuring influence, computing power indices through the direct formulas in Section~\ref{l_sect_prelim} is intractable, as one must examine all possible subsets or all permutations of the neurons. We approximate power indices via a Monte-Carlo method (akin to some existing methods~\cite{castro2009polynomial,Bachrach2010approximating,maleki2013bounding,michalak2013efficient,vstrumbelj2014explaining}), by averaging the marginal contributions over a fixed-size sample of randomly sampled coalitions. We allow biasing the procedure with a parameter $d$ so as to generate coalitions that retain approximately a proportion $d$ of the full model's neurons. 

Consider estimating the power index of player $a_i$. Given a sample $C_1, C_2, \ldots, C_k$ where $C_j \sim R_{\frac{1}{2}}$ (i.e. each coalition is randomly generated by flipping a fair coin per agent to decide whether they occur in it or not), an estimator for $a_i$'s Banzhaf index is: 
$\frac{1}{k} \sum_{j=1}^k \nu(C_j \cup \{ a_i \}) - \nu(C_j \setminus \{ a_i \})$. Similarly, to estimate $a_i$'s $t$-biased Banzhaf index $\beta_t(a_i)$ we can sample $C_1, C_2, \ldots, C_k$ where $C_j \sim R_t$ and apply the same formula (the only difference is we flip a {\it biased} coin for each agent).
Denoting by $\pi_j \sim R_{\Pi}$ the sampling of a permutation $\pi_j$ uniformly at random from the set of all agent permutations, an estimator for $a_i$'s Shapley value is $\frac{1}{k} \sum_{j=1}^k \nu(b(a_i, \pi_j) \cup \{ a_i \}) - \nu(b(a_i, \pi_j))$. We denote the procedure for estimating the power index of the agents as PIE$(\nu)$ where PIE stands for Power Index Estimation. Algorithm~\ref{alg:biased_banzhaf} describes PIE for the $t$-biased Banzhaf index $\beta_t$. Adapting this method to compute the Shapley value or Banzhaf index is trivial. \footnote{In order to reduce the variance in the Monte-Carlo estimates, the sampling of the coalitions can performed by first sampling a large set of random coalitions $C_1, \ldots C_k$ and their corresponding values $\nu(C_1), \ldots, \nu(C_k)$. Then, we estimate the power index of a player $i$ by subtracting the average value of the coalitions in the set that do not contain $i$ from the average value of coalitions including the player. A disadvantage of this approach is that the estimates of the power indices of the players are no longer independent variables. }

We note the PIE method can be easily modified to take into account an earlier decision to always include a certain subset $I$ neurons in the network, by changing the sampling procedure for the coalitions $C_1, \ldots, C_k$ to always include a certain neuron subset $I$. Alternatively we can reflect the assumption that a certain subset $I$ of neurons never occurs in the network by changing the sampling procedure to always exclude neurons in $I$. Further, we can compute the power indices of the neurons in a specific layer $L$ and assuming that all neurons in the other layers are not eliminated by always including these neurons not in $L$ in the sampled coalitions. 

\begin{algorithm}
  \begin{algorithmic}[1]
  \caption{Estimating the $t$-biased Banzhaf index $\beta_t$ for player $a_i$}
  \label{alg:biased_banzhaf}
    \Procedure{PIE}{$i$, $t$, $\nu$}
      \For{$j=1$ to $k$}
          \State $C_j \gets \emptyset$ 
          \For{player $x$ in $N \setminus \{ a_i \}$}
            \State $r \gets \mathcal{U}(0, 1)$                        \
            \If{$r < t$} \Comment{Include players with probability $t$}
                \State $C_j \gets C_j \cup \{ x \}$
            \EndIf
        \EndFor
      \EndFor
      \State \textbf{Return} $\frac{1}{k} \sum_{j=1}^k \nu(C_j \cup \{ a_i \}) - \nu(C_j \setminus \{ a_i \})$
    \EndProcedure
  \end{algorithmic}
\end{algorithm}

The biased power index $\beta_t$ reflects the relative importance of neurons under the assumption that they would be added to a random team of neurons of size $t$ in proportion to the original network's size. Assuming a parameter $d$ (selected by the MCUE procedure), let $\beta_d(i)$ be the power index estimation for neuron $i$ using the PIE procedure above. Denote the vector of all power indices as $\phi = (\phi_1, \phi_2, \ldots, \phi_n)$ where $\phi_i = \text{PIE}(i, d, \nu)$. A high value of $\phi_i$ indicates that the neuron $i$ is critical in many random sub-networks, whereas a low value of $\phi_i$ indicates that the neuron $i$ has relatively little impact when added to such a random sub-network and is a good candidate for pruning. 

\paragraph{Pruning Neurons Based on Power Indices} \label{sec:pruning_method}
We consider several methods for deciding which neurons to prune based on the power indices. The simplest methods are {\it non-iterative} and are based on a single estimation of the power indices. A more elaborate approach is {\it iterative computation}, that performs multiple iterations where in each iteration we re-estimate power indices based on the decisions made in the previous iteration. While iterative pruning can be more precise, it is also much more computationally demanding.   


The simplest method is {\bf Top-n Pruning}, a non-iterative procedure where we estimate the power indices and select the $r$ neurons with the heighest power indices (where $r$ is the target pruned network size). 
The {\bf Iterated Pruning} method is an iterative procedure, where we start with the full network and initial power index estimators. In each iteration we prune the $d$ neurons with lowest power indices, and then recompute power indices under the assumption that the pruned neurons are never included in any of the sampled coalition. By applying $\frac{n-r}{d}$ such iterations we eliminate $n-r$ neurons of the network, achieving a pruned network size of $r$. The {\bf Iterated Building} procedure is similar, except that we start with the empty network an initial power index estimator, and perform multiple iterations where in each iteration we add the $d$ neurons of highest power indices, then recompute power indices under the assumption that the retained neurons are always included. 

\section{Empirical Evaluation}
\label{sec:empirical}
We empirically evaluate the performance of GTAP and contrast it to multiple baselines~\cite{frankle2019lottery,blalock2020state}, by pruning both feedforward neural networks and convolutional neural networks. We carry our evaluation on two image classification datasets, MNIST~\cite{lecun1998lenet} and Tiny ImageNet~\cite{le2015tiny}, and on two natural language processing datasets, one for topic classification~\cite{TwitterFinancialNewsTopic2022} and one for social media text emotion classification~\cite{saravia-etal-2018-carer}. The neural architectures we prune are standard feedforward neural networks for natural language processing tasks (as well as LeNet-300-100~\cite{lecun1998lenet} for image classification), and the convolutional LeNet5~\cite{lecun1998lenet} and AlexNet~\cite{krizhevsky2012alexnet}. We note that the LeNet-300-100 and LeNet5 architectures are known to have low-size winning tickets~\cite{frankle2019lottery}. All networks were trained from random weight initialization, allowing us to perform a clean slate retraining of the networks to verify the winning ticket's success. For a full overview of the training procedure and corresponding training parameters, see 
Appendix A.



{\bf Experiments:}
Our empirical analysis is based on applying the GTAP method for power index based pruning using the various possible power indices (Shapley, Banzhaf and Biased Banzhaf with a parameter $d$ chosen using the uncertainty estimation). We compare to multiple baselines, such as Magnitude Pruning~\cite{frankle2019lottery} which prunes the weights with the lowest absolute values (anywhere in the entire neural network) or Gradient Pruning~\cite{blalock2020state} which prunes neurons with the lowest product of weight and gradient. 


For LeNet-300-100 all layers besides the output layer were pruned, whereas for LeNet5 the convolutional layers were not pruned to limit the complexity of experiments. Instead, the output features of the last convolutional layer were treated as input neurons, with pruning applied to this fully connected part of the network. \footnote{For AlexNet a dominant portion of layers are convolutional layers and so both convolutional and fully connected layers were pruned.}


We discuss the results of the uncertainty estimation in Section~\ref{sect:empirical_uncertainty} and outcomes of the neural pruning procedures in Sections~\ref{sect:empirical_gtap}.

\subsection{Uncertainty Estimation} 
\label{sect:empirical_uncertainty}
To establish the biasing factor for the power indices and the pruning process, we first obtain the uncertainty bands for a given \textit{trained} model and dataset. 

\begin{figure}[h!b]
    \centering
    \begin{subfigure}[b]{0.32\linewidth}
         \centering
         \includegraphics[width=\linewidth]{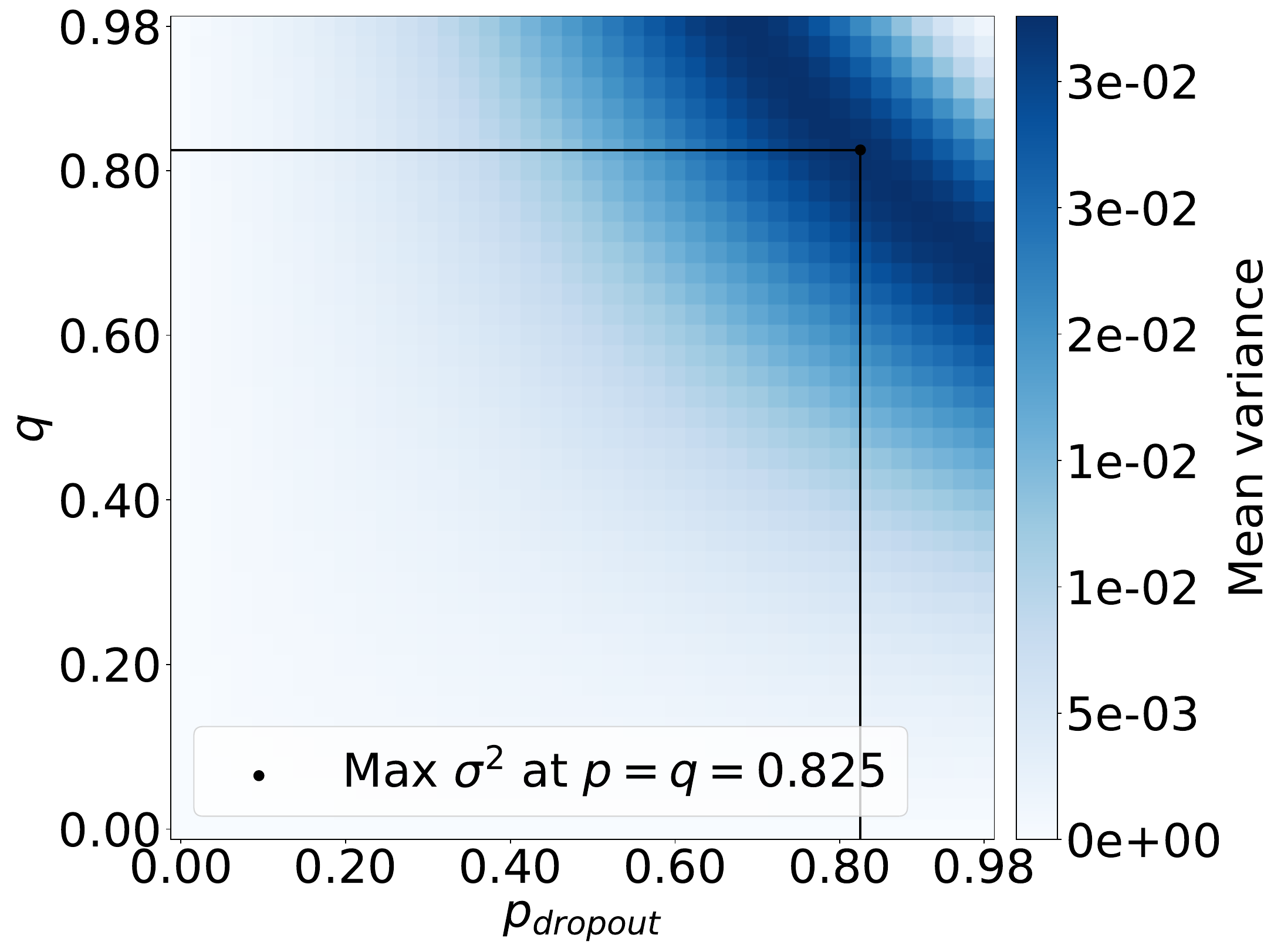}
         \caption{LeNet-300-100}\label{fig:lenet3_ue}
     \end{subfigure}
     \begin{subfigure}[b]{0.32\linewidth}
         \centering
         \includegraphics[width=\linewidth]{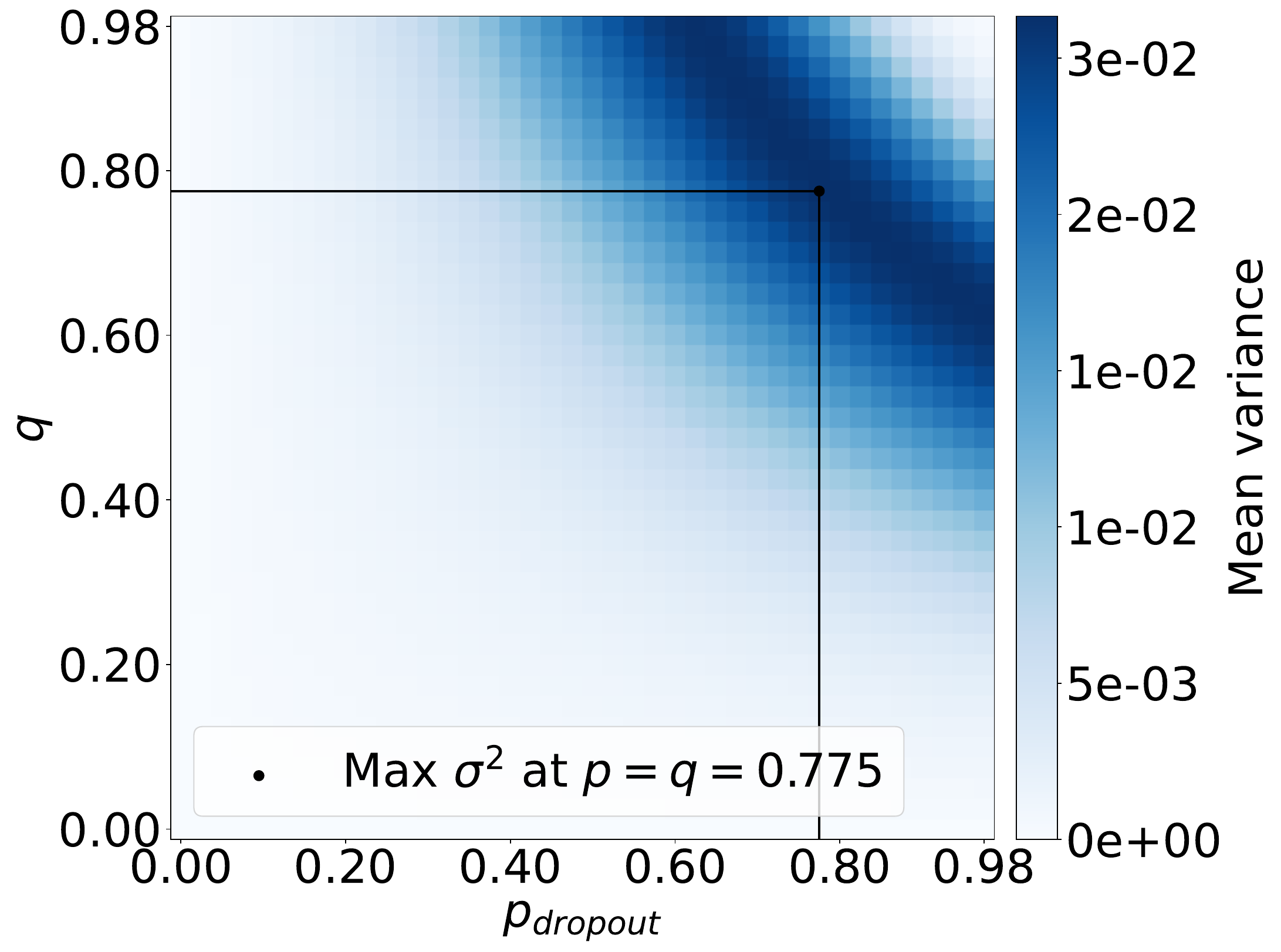}
         \caption{LeNet5}\label{fig:lenet5_ue_mc}
     \end{subfigure}
    \begin{subfigure}[b]{0.32\linewidth}
         \centering
         \includegraphics[width=\linewidth]{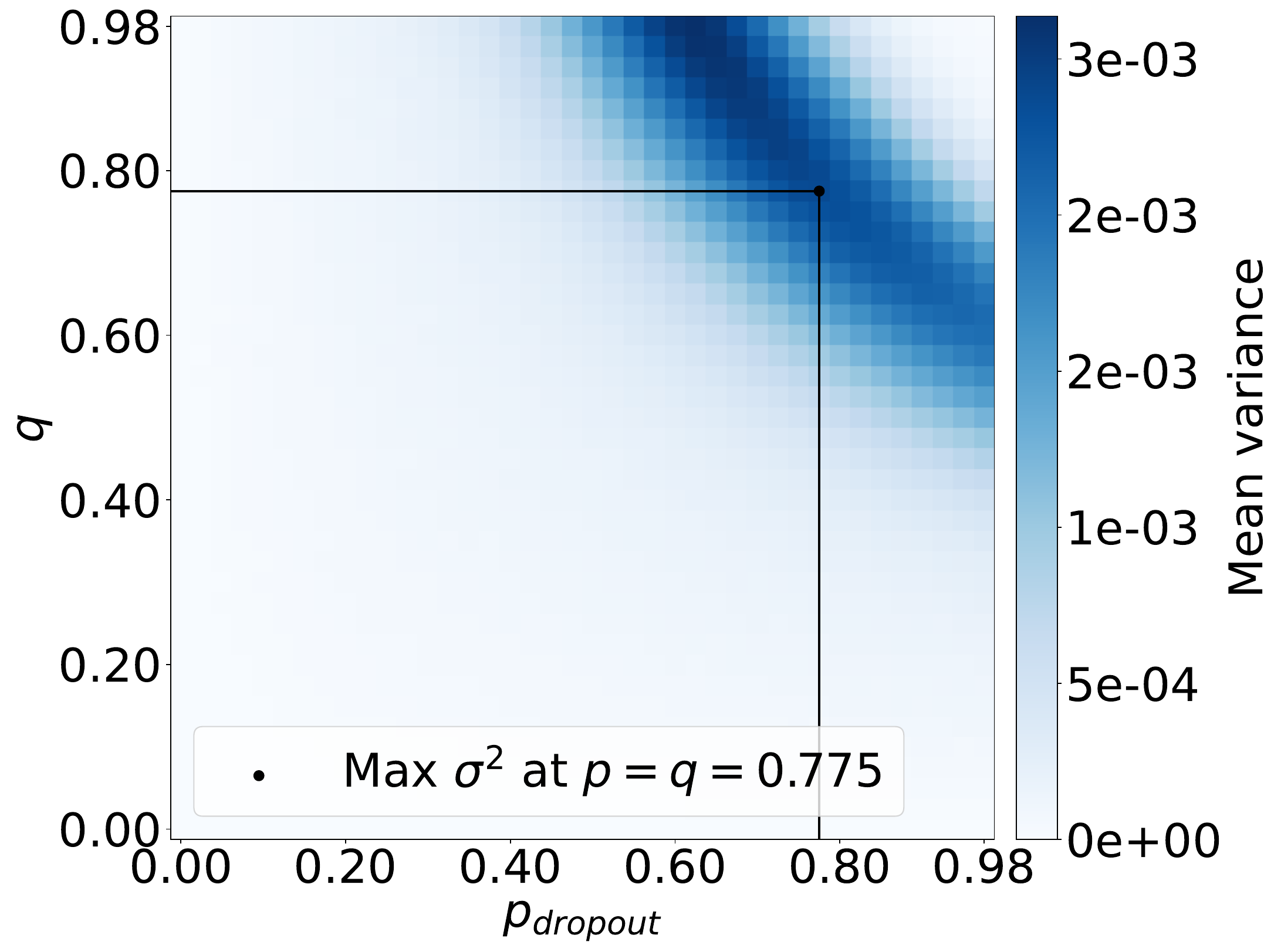}
         \caption{LeNet5 Binary}\label{fig:lenet5_ue_sc}
     \end{subfigure}
     
        \caption{Uncertainty estimation of the LeNet models with the MNIST dataset. Figure (a) and (b) show the model uncertainty for multi-class prediction, whereas Figure (c) shows the variance for the LeNet5 model trained for binary of just two classes. }\label{fig:lenet_ue}
\end{figure}

\subsubsection{LeNet}  
\label{sec:empirical_lenet}
Figure~\ref{fig:lenet_ue} shows the uncertainty bands for different LeNet architectures. We find that the variance stays at a relatively low level throughout the lower half of the $p/q$ values. The ``band'' of increased uncertainty is observed in the upper quadrant of the $p/q$ values, with the maximum variance on the diagonal at 0.825 and 0.775 for LeNet-300-100 and LeNet5 respectively. This suggests that the ticket network size of the LeNet-300-100 model is slightly smaller (0.175) as compared to the LeNet5 model (0.225). 

As a secondary test, we compared  the full 10-class MNIST task, shown in Figure~\ref{fig:lenet5_ue_mc}, to a binary classification task for digits 0 and 1. Though the band's location on the $p/q$ spectrum is the same for both tasks, the binary classification is narrower. Whereas the uncertainty bands for the LeNet-300-100 and the multi-class LeNet5 model are approximately symmetric along the diagonal, the binary classification model shows a gradual uncertainty increase from the bottom right to the top left of  band. Intuitively, this relates to the uncertainty difference between the two classes~\cite{kachman2019novel} where zero has more symmetry to the class and is easier to detect.

\subsection{Game Theoretic Pruning}
\label{sect:empirical_gtap}

We examine the ability of GTAP to prune neural networks while keeping the accuracy as high as possible. We examine ``compression curves'', where the x-axis reflects the target size for the pruned network, and the y-axis reflects the accuracy of the pruned model. Better pruning methods have curves that are higher for a wide range of pruned network sizes. 

\subsubsection{LeNet5} 
\label{sec:empirical_gtap_lenet}
\begin{figure}[h!t]
    \centering
    \begin{subfigure}[b]{0.80\linewidth}
         \centering
         \includegraphics[width=\linewidth]{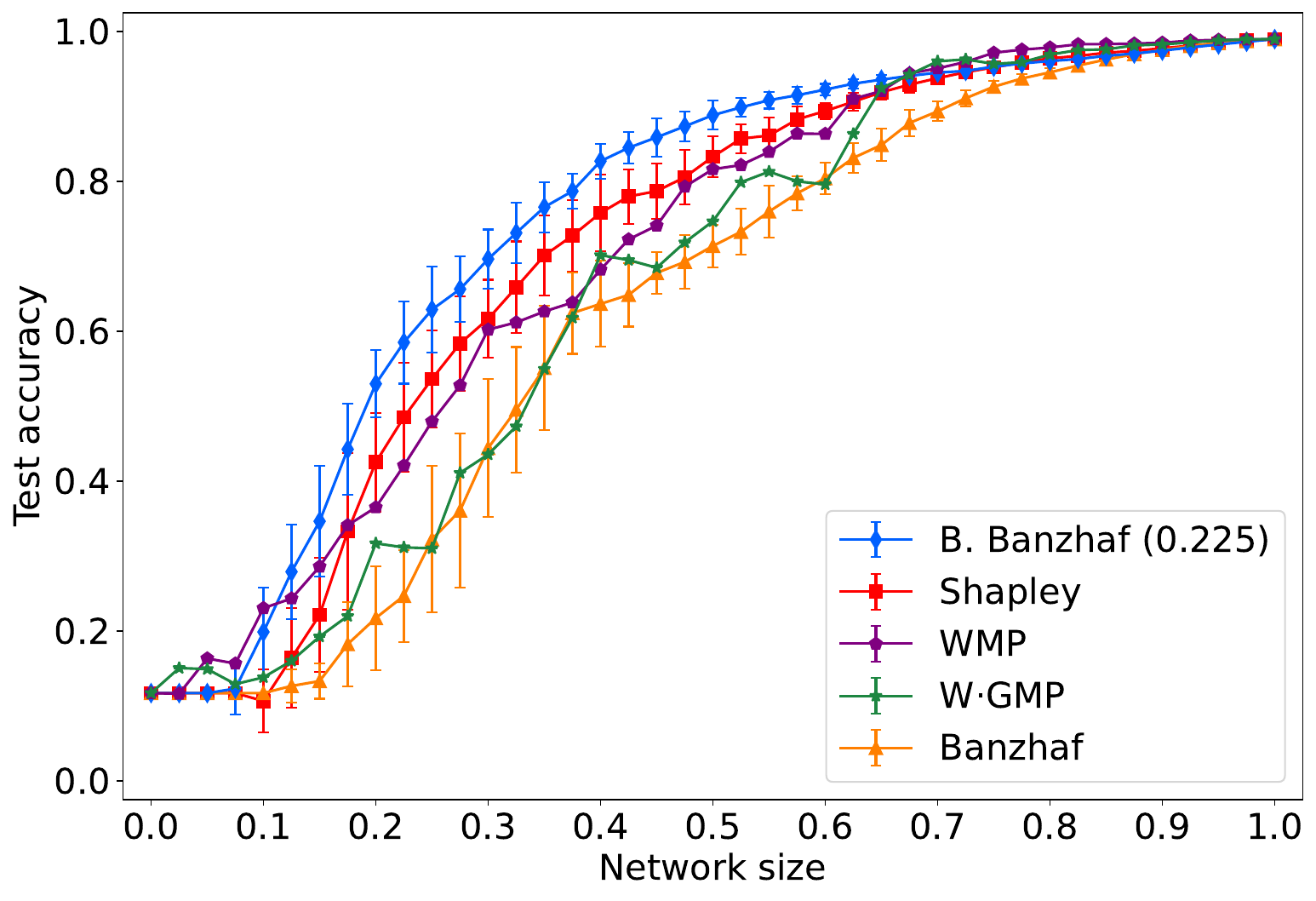}
         \caption{Top-n Pruning}
         \label{fig:pi_comp_topn}
     \end{subfigure}
     \begin{subfigure}[b]{0.80\linewidth}
         \centering
         \includegraphics[width=\linewidth]{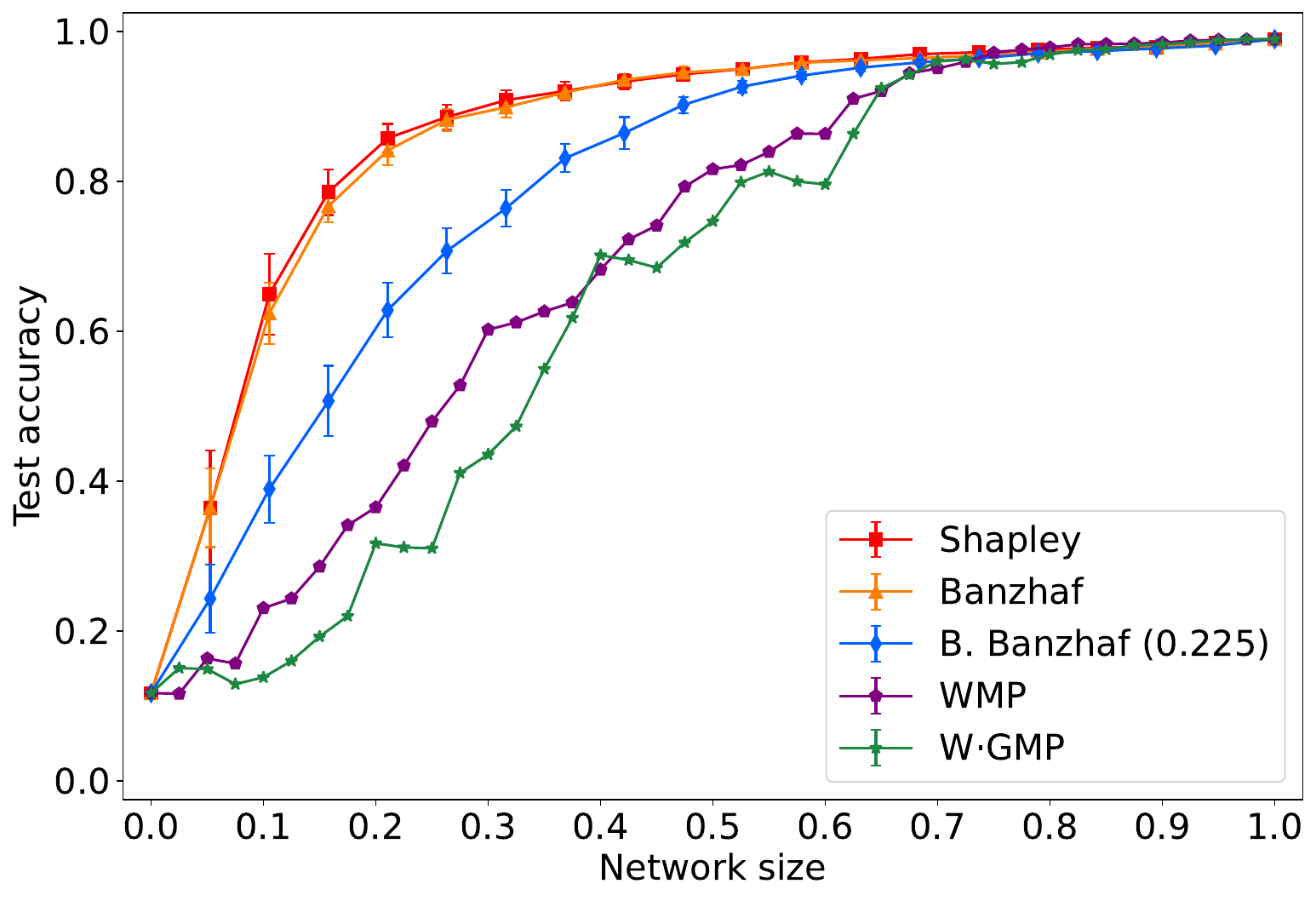}
         \caption{Iterated Pruning}
         \label{fig:pi_comp_tip}
     \end{subfigure}
        \caption{Comparing GTAP, Weight Magnitude Pruning ($WMP$) and Weight Gradient Magnitude Pruning ($W\cdot GMP$) on LeNet5.  Figure \ref{fig:pi_comp_topn} shows   Top-n Pruning, and Figure \ref{fig:pi_comp_tip} Iterated Pruning. 
        \label{fig:pi_comp}}
\end{figure}

Using the appropriate bias parameter $d$, that maximizes the uncertainty as shown in Figure~\ref{fig:lenet_ue}, we applied a both Top-n and Iterated Pruning for LeNet5 with MNIST, using the $d$-biased Banzhaf index, as well as the plain Banzhaf index and Shapely value. Figure~\ref{fig:pi_comp} compares the performance under these indices. 

Figure~\ref{fig:pi_comp} indicates a significant improvement in the performance of our GTAP method over the baselines (for all ranges of pruned network size). \footnote{Similar trends hold for LeNet-300-100.}
We note that Top-n Pruning exhibits higher improvement over the baseline, indicating that a one-shot selection of the highest power indices does not fully capture the importance of neuron interactions. 

As Figure~\ref{fig:pi_comp} shows, the Banzhaf index underperforms when sampling with the $d=\frac{1}{2}$ bias, but exhibits improved performance when $d$ is set to the critical network size estimated using the uncertainty bands (MCUE). The Shapley value exhibits performance in between random subset selection and the biased Banzhaf index. The detrimental performance of Banzhaf at $\frac{1}{2}$ seems to indicate that coalitions of size around $\frac{1}{2} |N|$ neurons are not indicative of the true influence of neurons on the prediction accuracy. The other results seem to indicate that coalitions around the critical network size are very informative, which benefits the biased Banzhaf approach.

For Iterated Pruning, the $t$-biased Banzhaf index approach performs significantly worse than the Shapley value and the default Banzhaf index. This suggests that approximating the Banzhaf power indices using coalitions around the critical network size is not as informative for small incremental pruning steps as uniform or $t=\frac{1}{2}$ biased sampling.

Figure~\ref{fig:pi_comp} shows that Iterative Pruning greatly outperforms Top-n Pruning, achieving a test accuracy of 0.871 at the critical network size, as compared to 0.585 for Top-n Pruning. Further, the variance between runs in Iterated Pruning at that size, $\sigma_{ITP} = 0.016$ is less than half that of the Top-n Pruning method, $\sigma_{TopN} = 0.055$.  The figure here shows only the strongest baselines, and Section B in the appendix contains a plot of the weaker baselines not plotted here. 


\begin{figure}[h!t]
    \centering
    \includegraphics[width=0.80\linewidth]{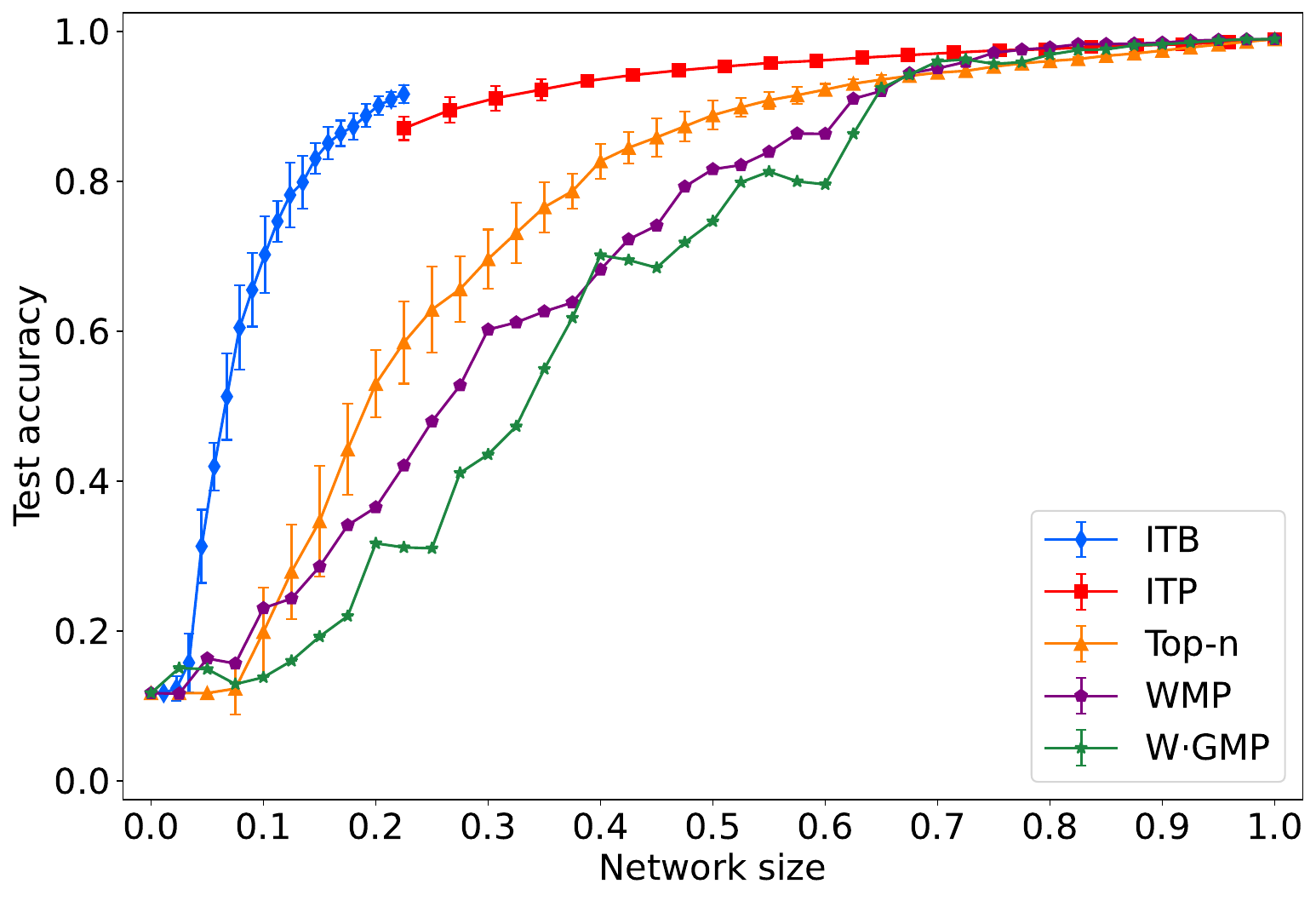}
    \caption{Comparison of the pruning methods on LeNet5. }\label{fig:prune_comp}
    \end{figure}

Comparing Iterated Building to Iterated Pruning, we find that Iterated Building outperforms Iterated Pruning for a similar compute budget (see Figure \ref{fig:prune_comp}). It achieves a test accuracy of 0.916 at the critical network size, compared to 0.871 for Iterated Pruning. Furthermore, the standard deviation of Iterated Building at that size is considerably lower ($\sigma_{ITB} = 0.012$ compared to $\sigma_{ITP} = 0.019$).

To summarize all the LeNet experiments, we find that GTAP achieves a significantly better compression-accuracy tradeoffs than all of the baselines (for all power indices and target network size). This shows that at least for these architectures and datasets, game theory can enable strong pruning methods.

\subsubsection{NLP Tasks} 
\label{sec:empirical_gtap_nlp}

Our results so far focused on image classification tasks. We now show that GTAP can also achieve good results in natural language processing. 
We consider two text classification tasks. One is a topic classification task~\cite{TwitterFinancialNewsTopic2022}, where the model examines a piece of text and must determine which news topic the text relates to. The second is an emotion classification task~\cite{saravia-etal-2018-carer}, where the model must determine which emotion characterizes the text (akin to sentiment classification but dealing with more fine-grained emotions). We apply a simple neural model to perform the classification; we consider a vocabulary $V$ of words, and represent each text as a binary term frequency vector (i.e. a text $t$ is represented as a vector $v^t$ of length $|V|$, where $v^t_i$ is $1$ if the $i$'th word in the vocabulary appears in the text and $v^t_i=0$ otherwise). Our model is a feedforward neural network with $k=3$ layers, each with $h=256$ neurons, applying a ReLU between each layer, and trained with the standard softmax cross-entropy loss. Following training the model, we apply our GTAP prunning method, and compare the performance to the weight based pruning baselines. 

\begin{figure}[h!t]
    \centering
    \begin{subfigure}[b]{1.00\linewidth}
         \centering
         \includegraphics[width=\linewidth]{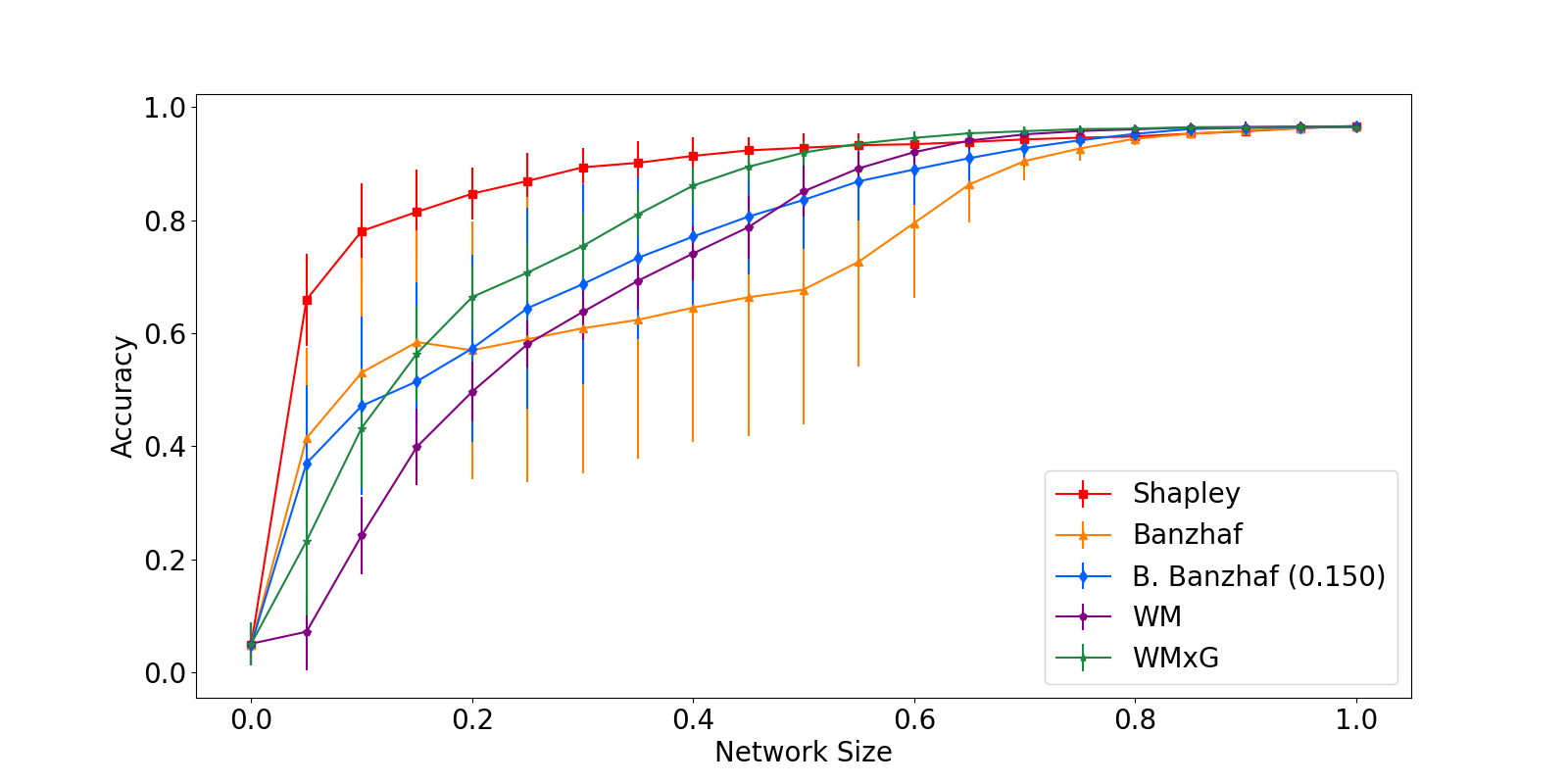}
         \caption{NLP: pruning a news topic classifier}
         \label{fig:twitter_news_topic}
     \end{subfigure}
     \begin{subfigure}[b]{1.0\linewidth}
         \centering
         \includegraphics[width=\linewidth]{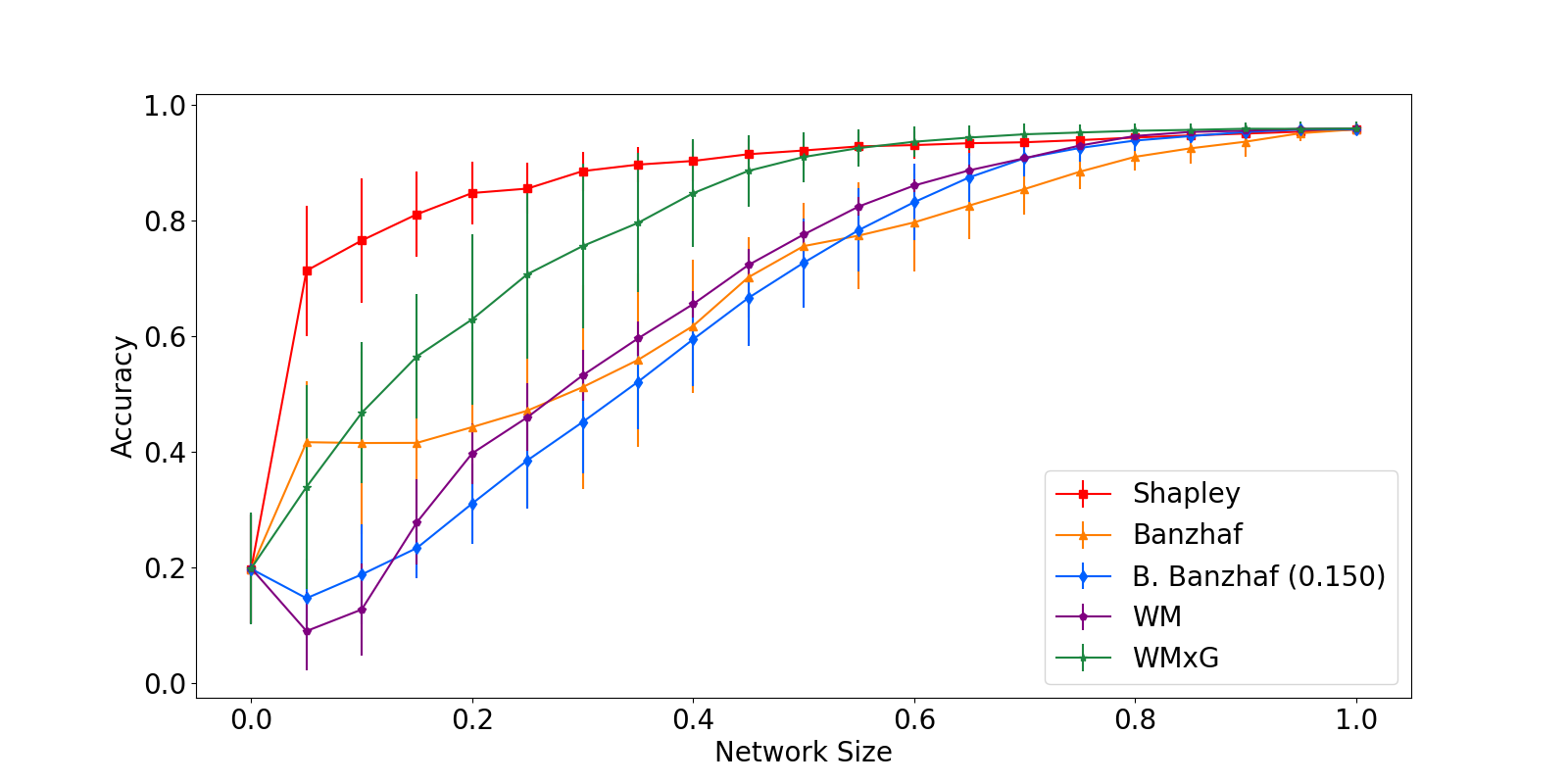}
         \caption{NLP: pruning a text emotion classifier}
         \label{fig:twitter_emotion}
     \end{subfigure}
        \caption{NLP tasks: Comparison of GTAP performance to baselines of Weight Magnitude Pruning ($WMP$) and Weight Gradient Magnitude Pruning ($W\cdot GMP$).  Figure \ref{fig:twitter_news_topic} shows results on the news topic classification task, and Figure \ref{fig:twitter_emotion} shows results for the text emotion classification task.
\label{fig:nlp_comp}}
\end{figure}

The results in Figure~\ref{fig:nlp_comp} indicate that our GTAP technique can be highly effective in the text modality as well, outperforming other pruning baselines. In this case, the Shapley based pruning outperforms the Banzhaf based one. Further research is needed to determine which power index is best suitable for other NLP datasets. Further, we pruned a very simple text classifier, and further work is needed to determine the best way to prune more advanced models, such as transformer based architectures. 

\subsection{Limitations and Scalability} 
\label{sec:LimitationsAlexnet}

We now discuss some limitations of our methods and analysis, describing experiments with the larger AlexNet~\cite{krizhevsky2012alexnet} architecture. 

First, we note that our experiments only consider convolutional neural networks and feedforward neural networks, so our methods might not have similar performance for other architectures (such as recurrent neural networks or transformers). 

Secondly, we used power indices to solve the cooperative game so as to identify important neurons. There is larger set of cooperative game solutions, such as the Core~\cite{gillies1953some}, Kernel~\cite{davis1965kernel} or Nucleolus~\cite{schmeidler1969nucleolus}. Applying such tools may yield a stronger performance. 

Finally, GTAP is computationally intensive, requiring high sample sizes both for computing the uncertainty bands (and bias parameter) and for computing the power indices of individual neurons. Large networks have a vast number of neurons, and hence GTAP requires computing power indices on very large games. Power indices are know to be computationally demanding~\cite{castro2009polynomial,Bachrach2010approximating,maleki2013bounding,michalak2013efficient}, and given a reasonable runtime, even strong approximation methods may return very inaccurate results. 

Indeed, Section~\ref{sec:alexnet_empirical} shows GTAP does not beat the baseline pruning algorithms on AlexNet~\cite{krizhevsky2012alexnet}. We conjecture that one reason for this are the inaccuracies in evaluating the power indices of the neurons due to the large size of the induced game. \footnote{We tackle some of this difficulty using layer-wise sampling of neurons. However, we have also experimented with a ``global'' sampling method, which achieved a lower performance. We found that under this global sampling, when correlating the resulting power index values with weight magnitude, the power indices were noisy. }
It is thus an important avenue for further research to see whether one can offer more accurate algorithms for solving cooperative games, that could be translated into a better performance in this case. 

\subsubsection{AlexNet Experiments}
\label{sec:alexnet_empirical}

\begin{figure}[h!t]
    \centering
    \begin{subfigure}[b]{0.32\linewidth}
         \centering
         \includegraphics[width=\linewidth]{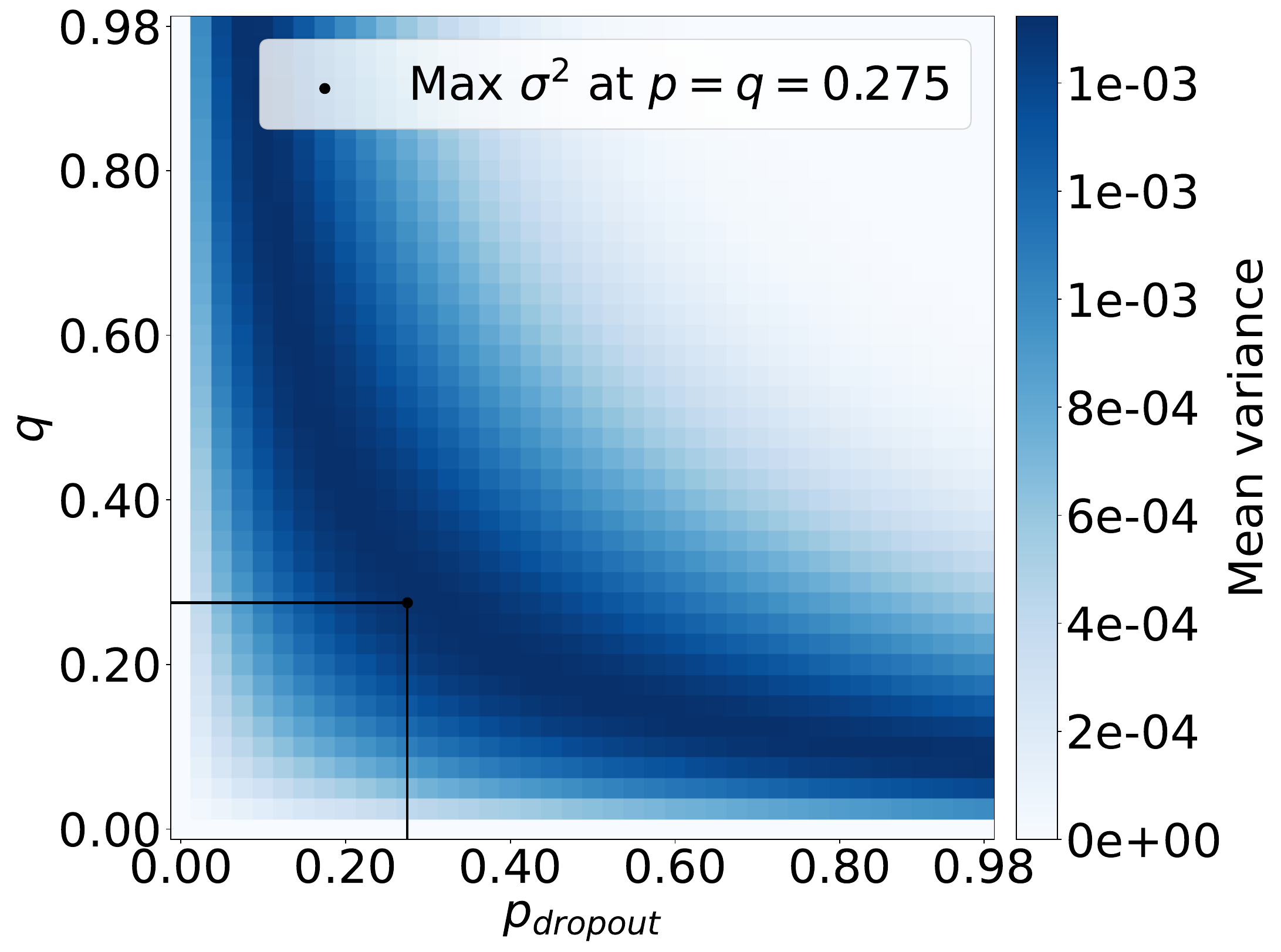}
         \caption{Full network}\label{fig:alexnet_ue_all}
     \end{subfigure}
     \begin{subfigure}[b]{0.32\linewidth}
         \centering
         \includegraphics[width=\linewidth]{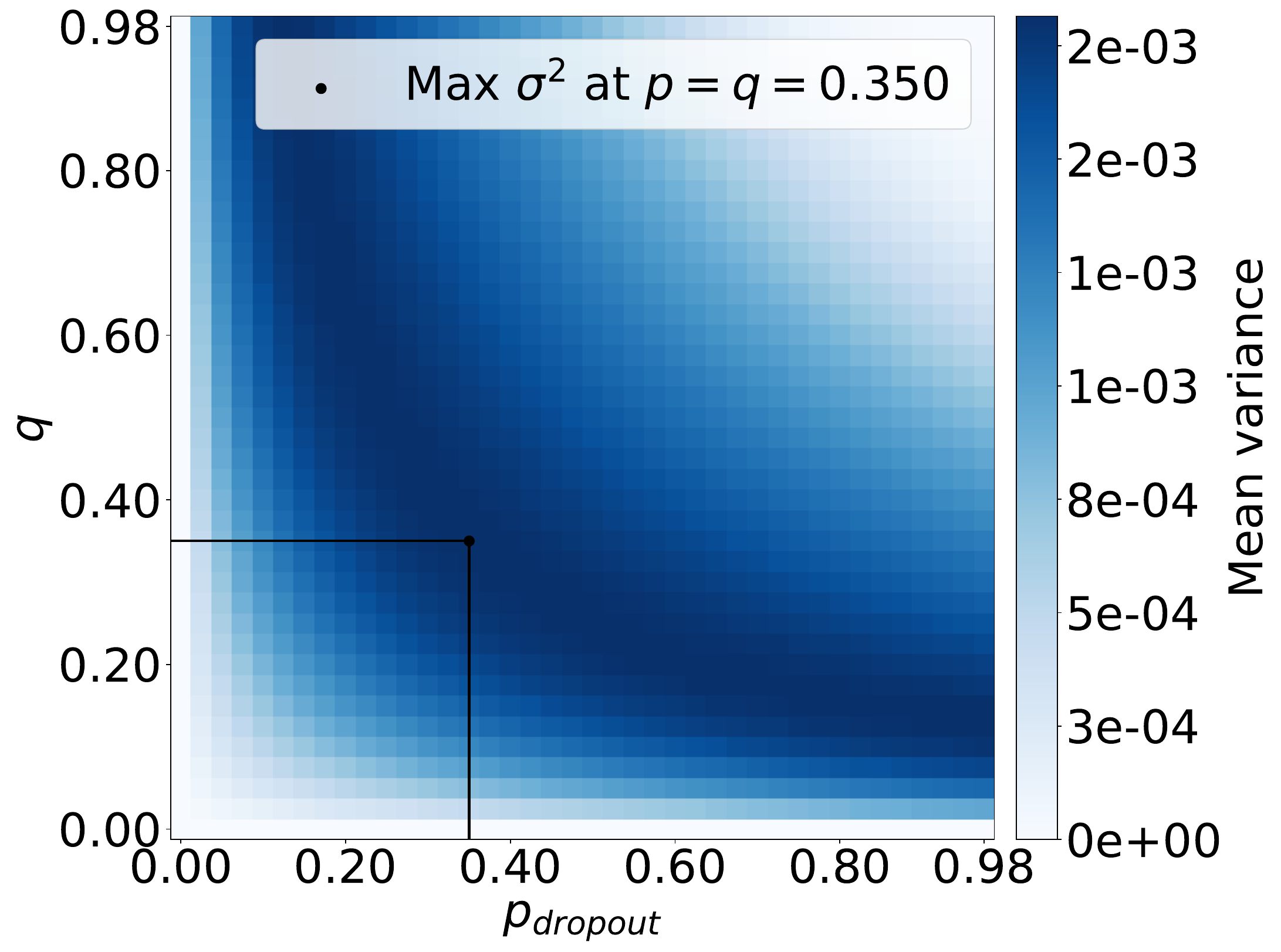}
         \caption{Conv. layers only}\label{fig:alexnet_ue_conv}
     \end{subfigure}
     \begin{subfigure}[b]{0.32\linewidth}
         \centering
         \includegraphics[width=\linewidth]{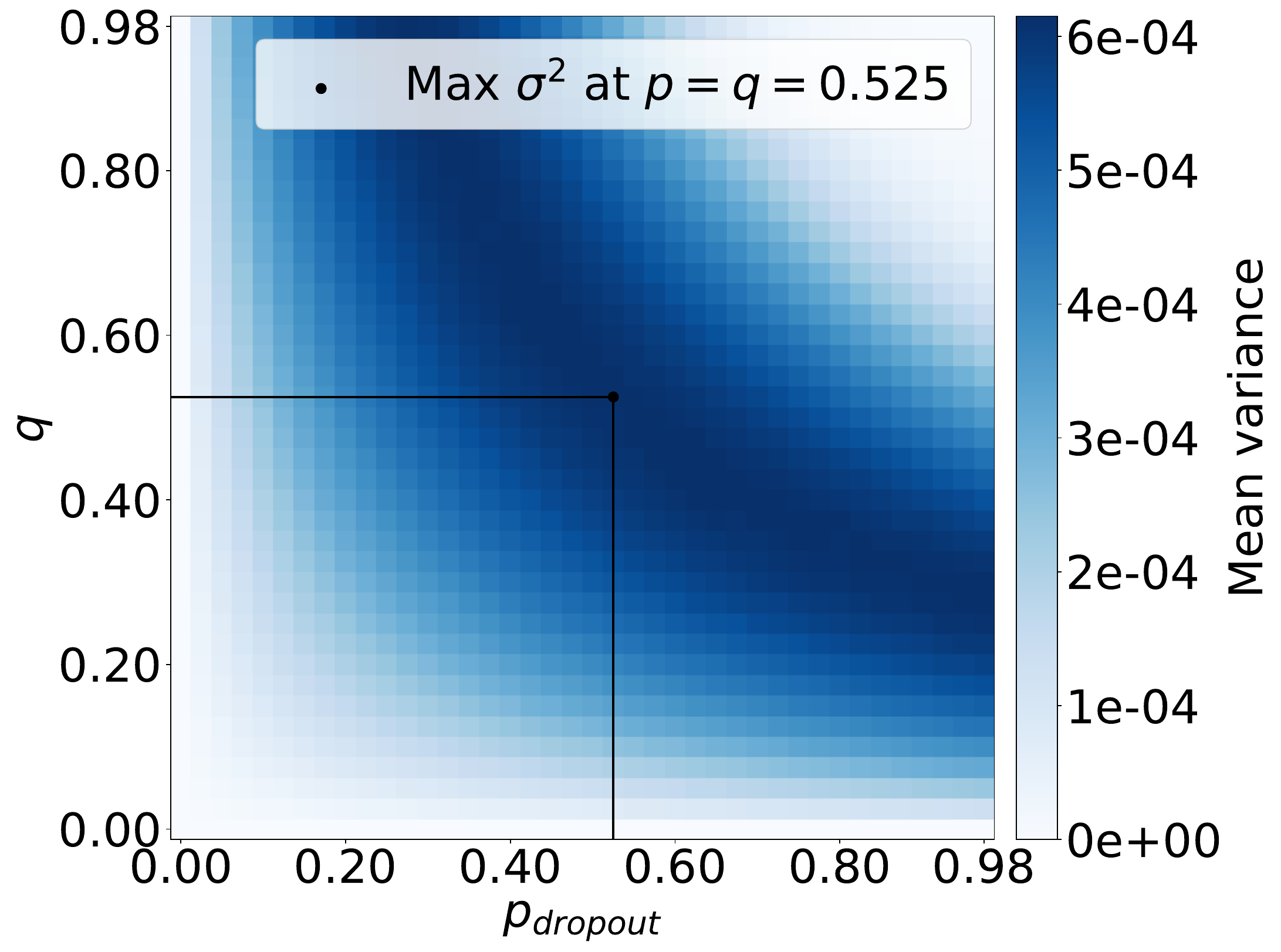}
         \caption{FC layers only}\label{fig:alexnet_ue_fc}
     \end{subfigure}
     
        \caption{Uncertainty estimation of the AlexNet model. Figure (a) shows the uncertainty band of the entire model. Figures (b) and (c) show the results of performing the uncertainty estimation by ablating only within the convolutional  (b) layers or the fully connected layers  (c) respectively.}\label{fig:alexnet_ue}
\end{figure}

We now present our results on AlexNet. Figure~\ref{fig:alexnet_ue} shows the uncertainty bands for AlexNet. The bands occur in a lower parameter regime, so the critical network size for AlexNet is likely larger than for the LeNet models. 
We also note that the full network only achieves an accuracy of 51.6\%, indicating a more difficult task.
We observe that the uncertainty bands differ significantly when quantifying the uncertainty of different types of layers. Figures (\ref{fig:alexnet_ue_conv}) and (\ref{fig:alexnet_ue_fc}) show  that the convolutional layers have a larger critical network size then a fully connected one, hinting that they play a different role in the networks functionality and the the winning tickets composition. 

\paragraph{GTAP on AlexNet:}
The large size of AlexNet makes it too big to estimate the power indices of individual neurons, forcing us to 
approximate power indices {\it for each layer} separately, given the assumption that the neurons in the other layers are always present. This enables us to rank neurons in each layer, but does not allow comparison across layers. Instead, we consider the baseline of Weight Magnitude Pruning (with target network size of $0.3$ of the entire network, where this pruning methods shows saturation); we examine the proportion of model parameters in each layer that are selected into the pruned network by this baseline. 

We determine which neurons GTAP selects by examining the layer based power indices, and for each layer select its respective proportion of neurons with the highest power indices. In other words, we use Weight Magnitude Pruning to determine how many neurons to select in each layer, and use GTAP to select the neurons of highest power indices in each layer. 

\begin{figure}[t!h]
    \centering
        \includegraphics[width=0.80\linewidth]{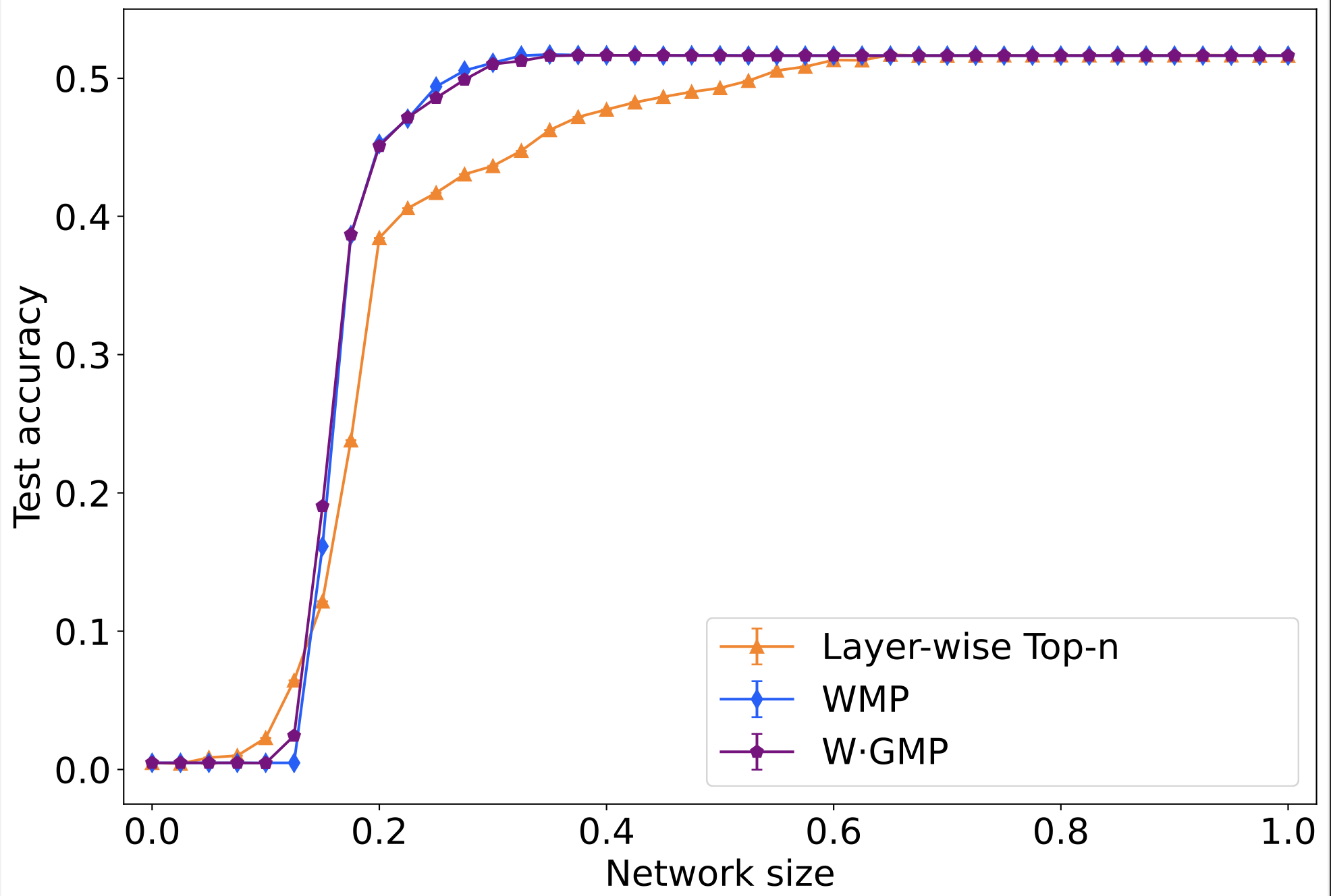}
        \caption{Comparison of the performace of GTAP with layer-wise Top-n Pruning and the baselines of Weight Magnitude Pruning ($WMP$) and Weight Gradient Magnitude Pruning $W\cdot GMP$ on the AlexNet model.}\label{fig:alexnet}
\end{figure}

The ``compression curves'' for AlexNet are shown in Figure~\ref{fig:alexnet}. As the figure shows, the Iterated Building results for AlexNet are very different from LeNet5, which showed a fast increase in test accuracy. 
For AlexNet, the increase starts at roughly 10\% of the players, and plateaus into an approximately linear increase in test accuracy around 20\% of the network size.
Figure~\ref{fig:alexnet} shows that for AlexNet, WMP and W-GMP slightly outperform GTAP. \footnote{In the case where the power index approximation is guided by the same data as evaluation, we do find a slight improvement over the baselines when retaining a fraction of 40\% to 60\% of the original network. However, this overfits the neuron selection to the test data, making an unfair comparison.} 

\begin{figure}[!ht]
    \centering
    \begin{subfigure}[t]{0.30\linewidth}
         \centering
         \includegraphics[width=\linewidth]{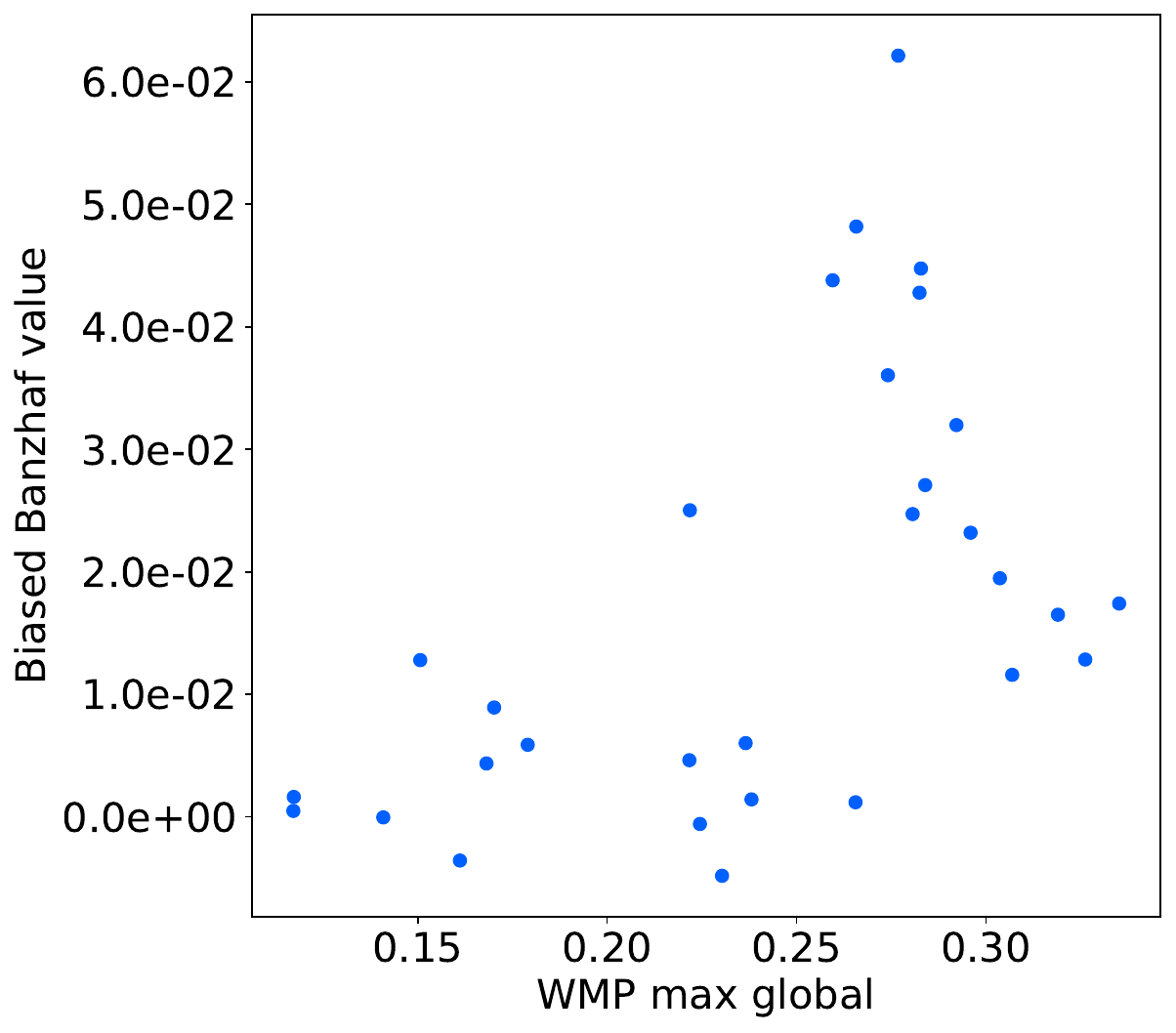}
         \caption{Conv1}\label{fig:corr1}
     \end{subfigure}
     \begin{subfigure}[t]{0.30\linewidth}
         \centering
         \includegraphics[width=\linewidth]{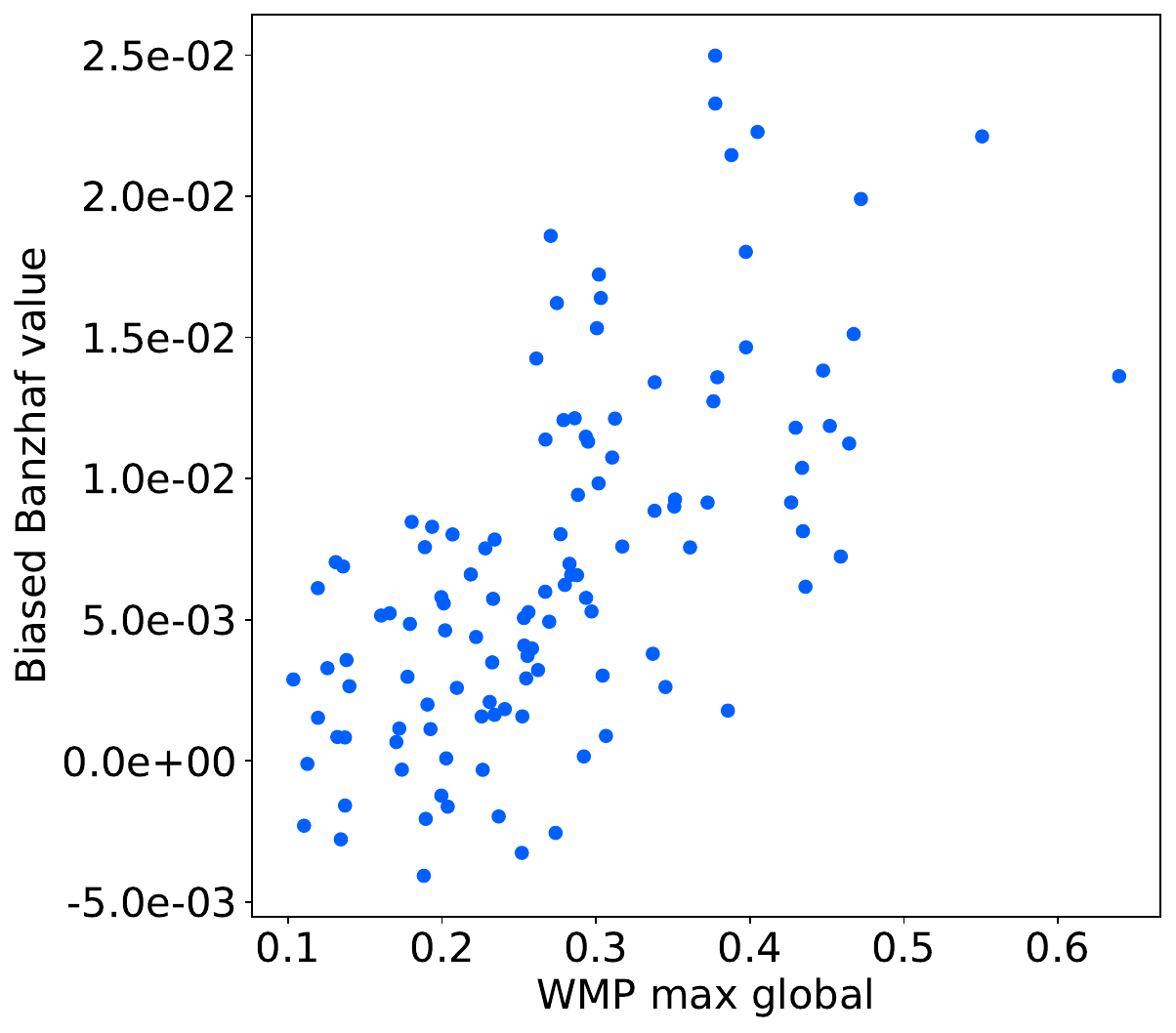}
         \caption{Conv2}\label{fig:corr2}
     \end{subfigure}
     \begin{subfigure}[t]{0.30\linewidth}
         \centering
         \includegraphics[width=\linewidth]{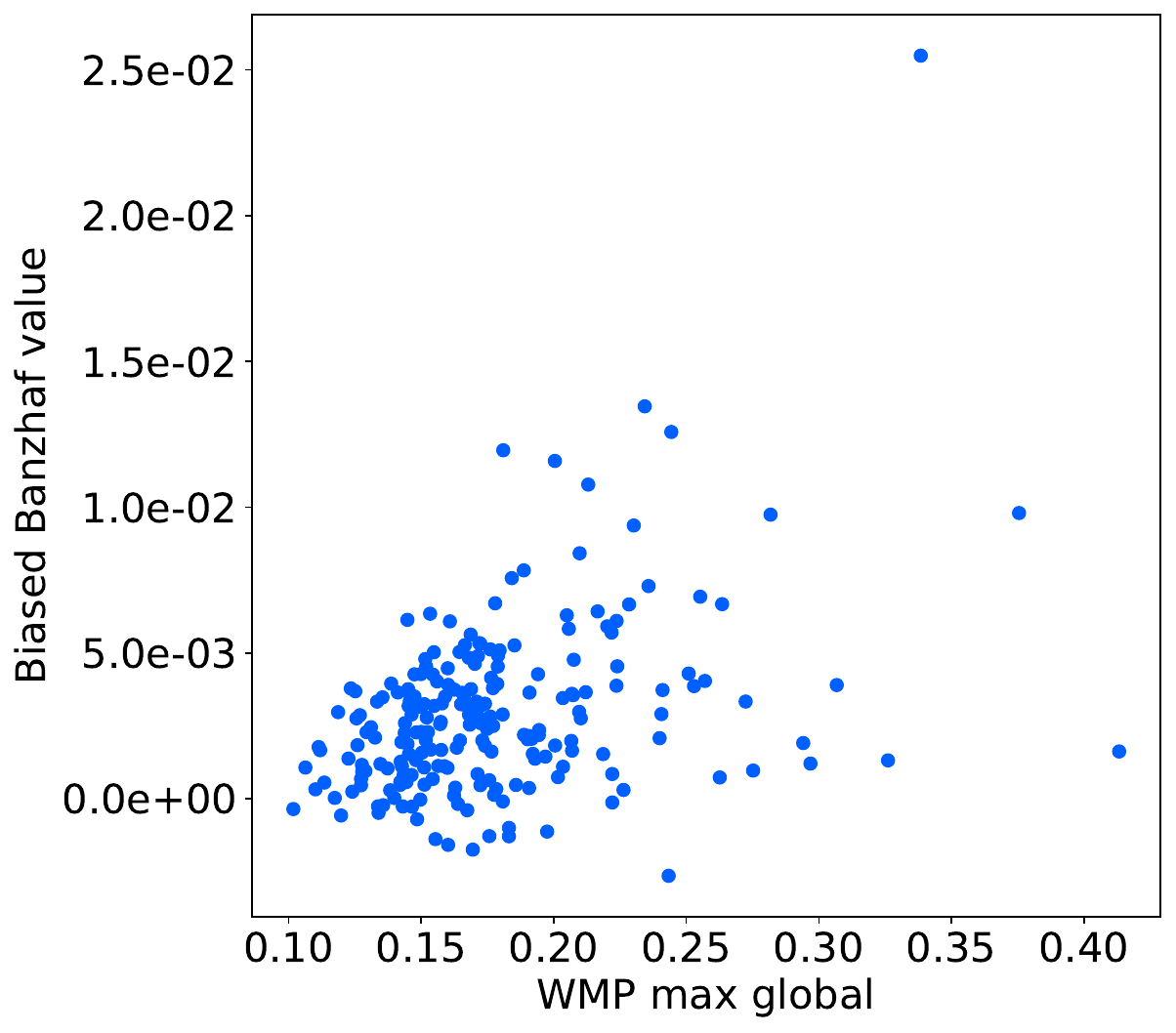}
         \caption{Conv3}\label{fig:corr3}
     \end{subfigure}\\
     \begin{subfigure}[b]{0.30\linewidth}
         \centering
         \includegraphics[width=\linewidth]{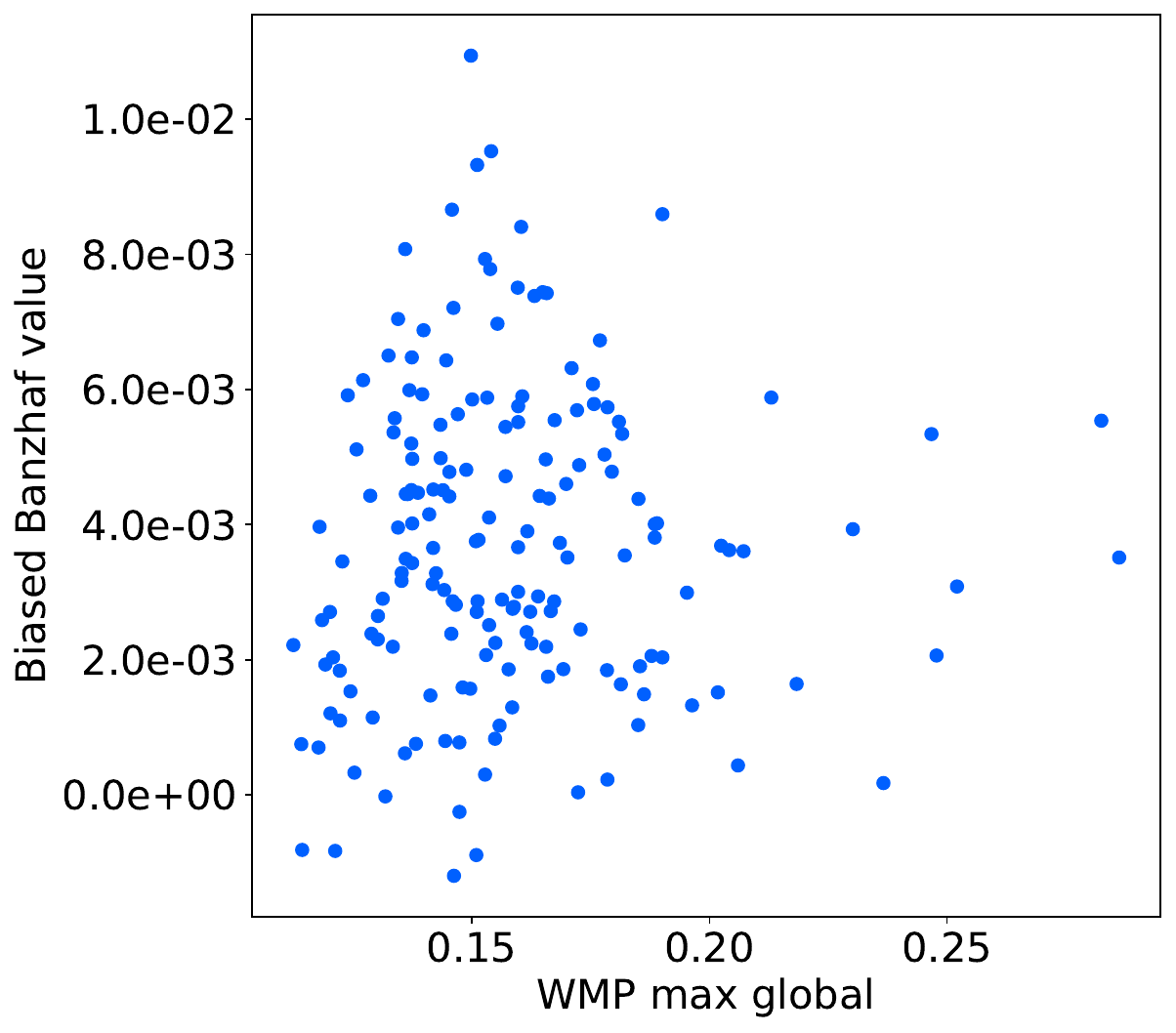}
         \caption{Conv4}\label{fig:corr4}
     \end{subfigure}
     \begin{subfigure}[b]{0.30\linewidth}
         \centering
         \includegraphics[width=\linewidth]{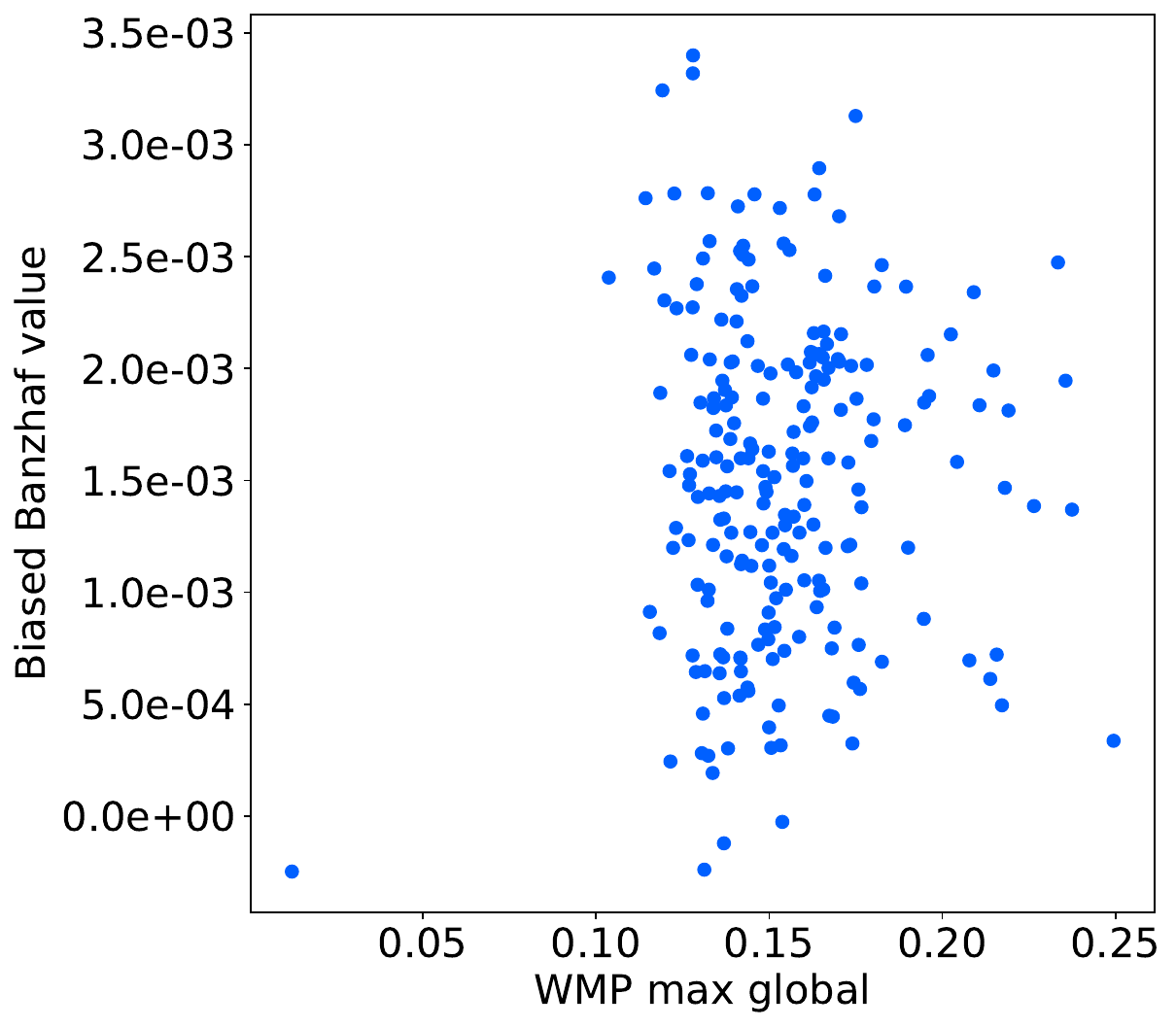}
         \caption{Conv5}\label{fig:corr5}
     \end{subfigure}
     \begin{subfigure}[b]{0.30\linewidth}
         \centering
         \includegraphics[width=\linewidth]{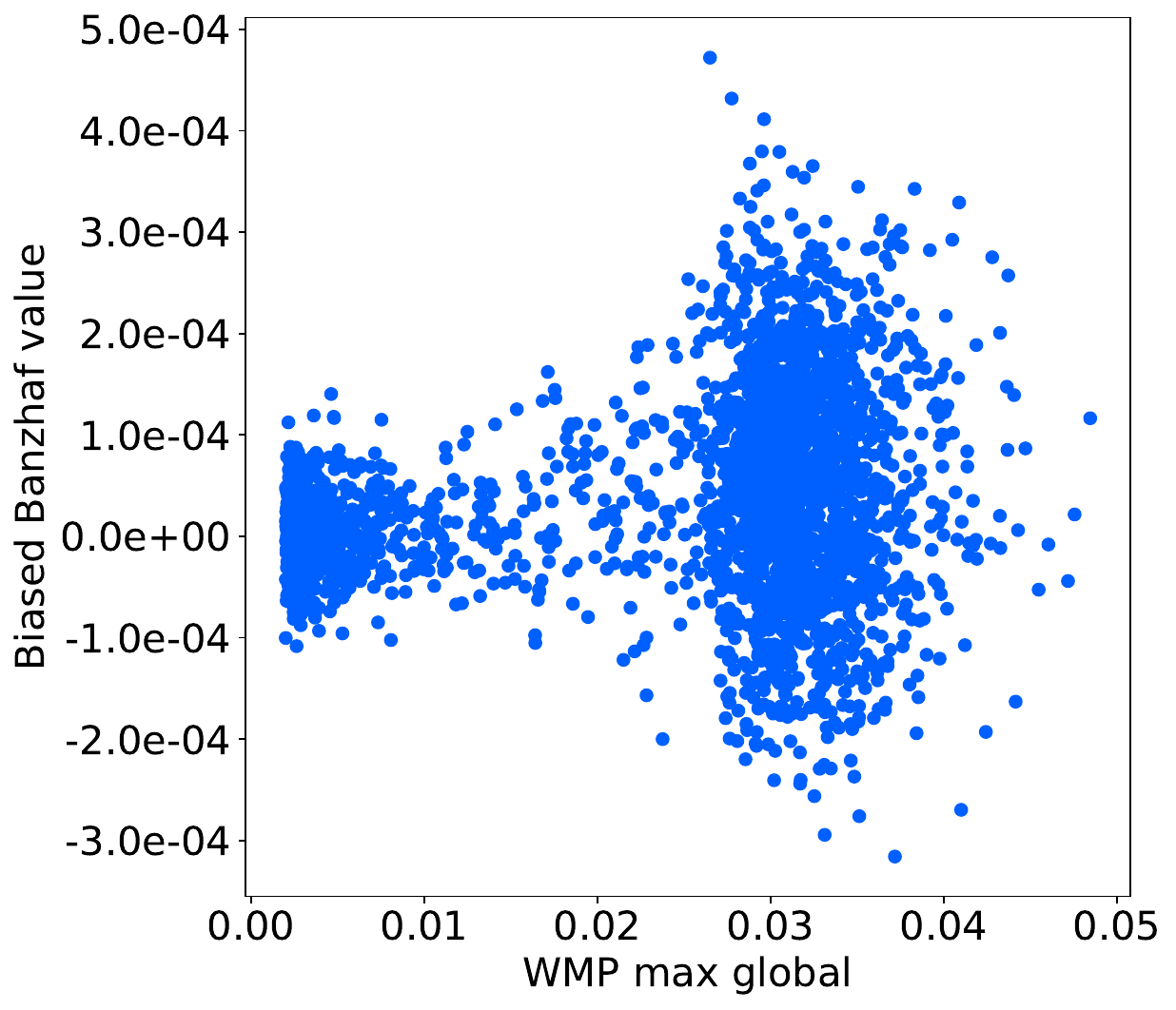}
         \caption{Fc1}\label{fig:corr6}
     \end{subfigure}\\
     \begin{subfigure}[b]{0.30\linewidth}
         \centering
         \includegraphics[width=\linewidth]{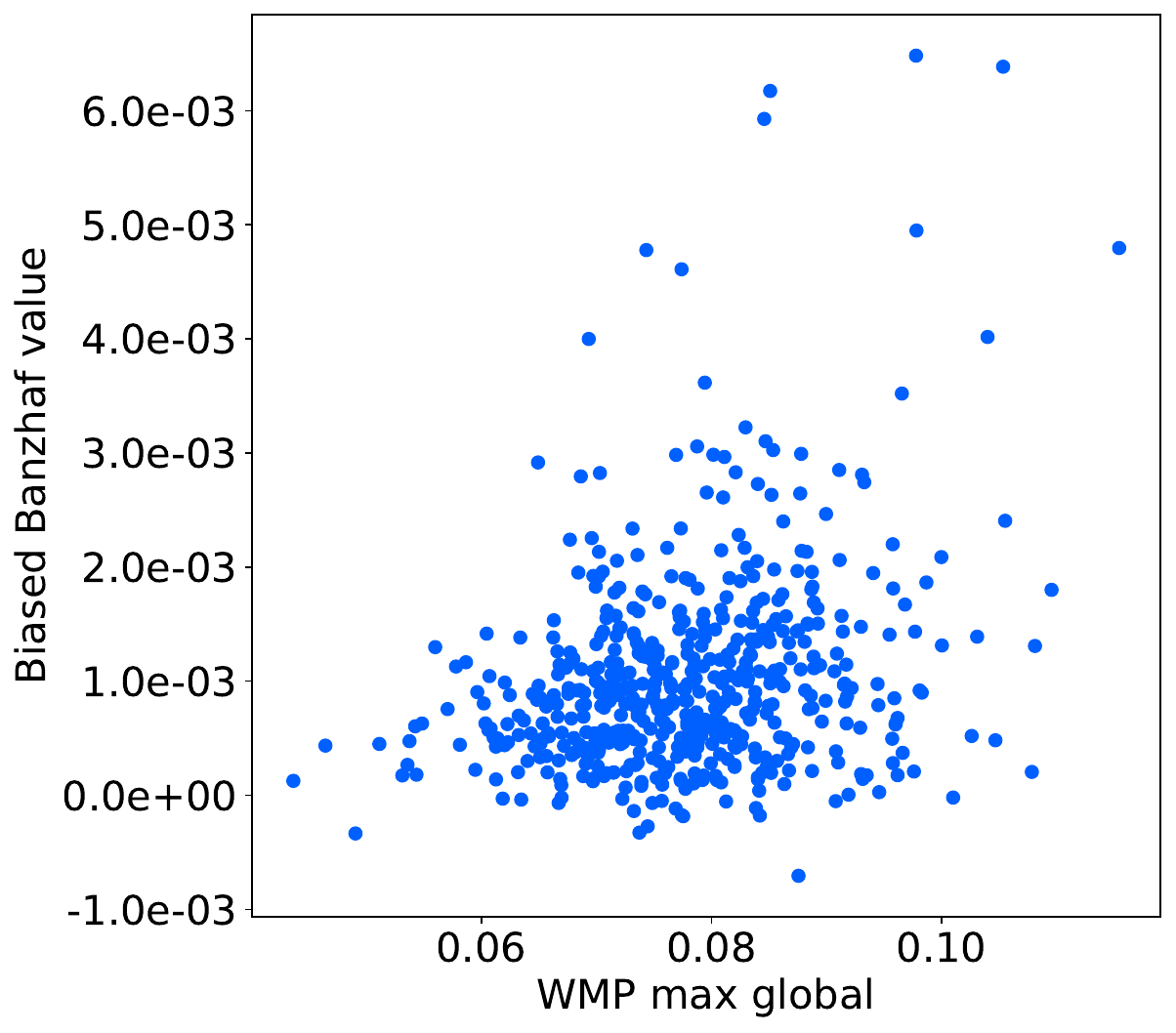}
         \caption{Fc2}\label{fig:corr7}
     \end{subfigure}
        \caption{Correlations plots between power indices in layer-wise Pruning and Weight Magnitude Pruning (Alexnet)}\label{fig:correlations}
\end{figure}

We also investigated whether GTAP and the WMP baseline select a similar sub-network when pruning AlexNet. Figure~\ref{fig:correlations} shows correlation plots between the power index of neurons (their importance ranking under GTAP) and their weight magnitude (their importance ranking under WMP). It visualizes the correlations between the GTAP and WMP rankings for neurons in each layer. \footnote{To ease the approximation of power indices, We have omitted neurons with extremely low weights, of below $2\times 10^{-3}$.  Hence, these neurons are not included in the figure.}

Figure~\ref{fig:correlations} indicates that correlation is strongest in the initial convolutional layers, with the latter convolutional layers and the fully connected layers showing reduced correlation. All of the figures indicate some correlation between the GTAP and WMP rankings, but also a very wide dispersion. The GTAP and WMP rankings are quite different, so despite having similar performance, these two methods construct quite different sub-networks. This means that potentially one could find a new pruning method that could combine the strengths of these two methods.  

\section{Discussion and Conclusions}
\label{sec:discussion}

We examined an application of game theory in improving the efficiency of deep learning architectures, and proposed a method called GTAP that prunes neural networks based on solution concepts for cooperative games. In particular, we applied power indices, including the Shapley value~\cite{Shapley1953} and Banzhaf index~\cite{BanzhafIII1964}, as well as a paremetrized biased version of the Banzhaf index. Our method selects a good bias parameter based on uncertainty estimation using a procedure akin to Dropout~\cite{srivastava2014dropout}.
We showed empirical results on two image classification datasets (MNIST~\cite{lecun1998lenet} and Tiny ImageNet~\cite{le2015tiny}) using multiple convolutional neural architectures, LeNet5~\cite{lecun1998lenet}  and AlexNet~\cite{krizhevsky2012alexnet}),
and two natural language processing datasets (topic classification~\cite{TwitterFinancialNewsTopic2022} and social media emotion classification~\cite{saravia-etal-2018-carer}).  

Our results indicate that game theoretic pruning can be a strong tool in reducing the size and compute requirement of neural network models, while retaining good predictive performance. We show that in many cases, GTAP outperforms existing strong baselines such as Weight Magnitude Pruning (WMP) and Weight Gradient Magnitude Pruning (W-GMP)~\cite{blalock2020state}. 

We note that GTAP offers several advantages as a pruning method. First, GTAP does not require retraining or finetuning the model during the pruning process (though it would be interesting to examine whether this could further improve  performance). Secondly, GTAP is highly parallelizable, as one can divide the sampling process over multiple machines. Finally, GTAP is model agnostic and can be extend to other applications, so long as the units to be pruned can be dropped from the computation (for example ensemble models or random forests). This may prove useful in machine learning methods where weight magnitudes do not occur or are inaccessible.

{\bf Future work:} several questions remain open for further research. First, can other game theoretic solutions as such other power indices~\cite{deegan1978new}, the core or similar solutions~\cite{gillies1953some,faigle2001computation,bachrach2018bounds} be used to further improve pruning? Second, could one find better approximation algorithms for power indices so as to achieve a higher pruning performance? \footnote{While power index computation is a hard problem even in restricted games~\cite{aziz2009algorithmic,aziz2009power,bachrach2007computing,elkind2009tractable,bachrach2013proof,xiao2022shapley,see2014cost}, there may be good approximation algorithms for the average case~\cite{covert2020improving,jethani2021fastshap}.}
Finally, how can one extend our methods to larger or more elaborate neural network architectures, such those driving large language models?

\clearpage
\bibliographystyle{abbrv}
\bibliography{library}
\clearpage

\end{document}